\documentclass[twoside,11pt]{article}

\usepackage{arxiv}

\usepackage{Definitions}
\usepackage{slashbox}
\usepackage[normalem]{ulem}
\usepackage{amssymb}
\usepackage{amsmath}
\usepackage{color}
\usepackage{times}

\usepackage{graphicx}
\usepackage{subcaption}
\usepackage{float}
\usepackage{booktabs}
\usepackage[hyphenbreaks]{breakurl}

\hypersetup{
    colorlinks=true,
    linkcolor=black,
    citecolor=black,
    filecolor=black,
    urlcolor=black,
}

\jmlrheading{1}{2019}{XX}{XX}{XX}{YY}{K. Kawaguchi, Y. Bengio, V. Verma \& L. Kaelbling}

\ShortHeadings{Analytical Learning Theory}{Kawaguchi, Bengio, Verma, and Kaelbling}
\firstpageno{1}
\usepackage{color}

\begin{document}

\title{Generalization in Machine Learning via Analytical Learning Theory} 
\author{\name Kenji Kawaguchi \email kawaguch@mit.edu\\
       \addr Massachusetts Institute of Technology\\
       77 Massachusetts Ave, Cambridge, MA 02139, USA\\
       \AND
       \name Yoshua Bengio  \email yoshua.bengio@umontreal.ca\\
       \addr University of Montreal\\
       2900 Edouard Montpetit Blvd, Montreal, QC H3T 1J4, Canada\\      
       \AND
       \name Vikas Verma  \email vikas.verma@aalto.fi\\
       \addr Aalto University\\
       Konemiehentie 2, Espoo, Finland\\      
       \AND
       \name Leslie Pack Kaelbling \email lpk@csail.mit.edu\\
       \addr Massachusetts Institute of Technology\\
       77 Massachusetts Ave, Cambridge, MA 02139, USA\\       
       }

\editor{XX}

\maketitle

\begin{abstract}
This paper introduces a novel measure-theoretic learning theory for machine learning that does not require statistical assumptions. Based on this theory, a new regularization method in deep learning is derived and shown to outperform previous methods in CIFAR-10, CIFAR-100, and SVHN. Moreover, the proposed theory provides theoretical justifications for one-shot learning, representation learning, deep learning, and curriculum learning as well as a family of practically
successful regularization methods. Unlike statistical learning theory, the proposed learning theory analyzes each problem instance individually via measure theory, rather than a set of problem instances via statistics. As a result, it provides different types of results and insights when compared to statistical learning theory.
\end{abstract}


\section{Introduction}

Statistical learning theory provides  tight and illuminating results under its assumptions and for its objectives (e.g., \citealt{vapnik1998statistical,mukherjee2006learning,mohri2012foundations}). As the training datasets are considered as random variables, statistical learning theory was initially more concerned with the study of  \textit{data-independent} bounds based on the capacity of the hypothesis space \citep{vapnik1998statistical}, or the classical  stability of learning algorithm \citep{bousquet2002stability}. Given the observations that these data-independent bounds could be overly pessimistic for a ``good'' \textit{training dataset}, \textit{data-dependent} bounds have also been  developed in statistical learning theory, such as  the \textit{luckiness framework} \citep{shawe1998structural,herbrich2002algorithmic}, \textit{empirical} Rademacher complexity of a hypothesis space \citep{koltchinskii2000rademacher,bartlett2002model}, and the robustness of learning algorithm \citep{xu2012robustness}.   

Along this line of reasoning, we notice that the previous bounds, including data dependent ones, can be   pessimistic for a ``good'' \textit{problem instance}, which is defined by a tuple of a true (unknown) measure, a training dataset and a learned model (see Section \ref{sec:basis} for further details). Accordingly, this paper  proposes a learning theory designed to be strongly dependent on each individual problem instance. To achieve this goal, we directly analyse the generalization gap (difference between expected error and training error) and datasets as  non-statistical  objects via measure theory. This is in contrast to the setting of statistical learning theory wherein these objects  are treated as random variables. 

The non-statistical nature of our proposed theory can be   of practical interest on its own merits. For example, the non-statistical nature  captures  well a  situation wherein a training dataset is specified and fixed first  (e.g., a UCL dataset, ImageNet, a medical image dataset, etc.), rather than remaining random with a certain  distribution. 
Once a dataset is actually specified, there is no randomness remaining over the dataset (although one can artificially create randomness via an empirical distribution). For example, \citet{zhang2016understanding} empirically observed that \textit{given a fixed (deterministic) dataset} (i.e., each of CIFAR10, ImageNet, and MNIST), test errors  can be  small despite  the large capacity of the hypothesis space and possible instability of the learning algorithm. Understanding and explaining this empirical observation has  become an active research area  \citep{arpit2017closer,krueger2017deep,hoffer2017train,wu2017towards,dziugaite2017computing,dinh2017sharp,bartlett2017spectrally,brutzkus2017sgd}.

For convenience within this paper, the proposed theory is called   \textit{analytical learning  theory}, due to its non-statistical and analytical nature. While the scope of  statistical learning theory covers both prior and posterior guarantees, analytical learning theory focuses on providing \textit{prior} insights  via \textit{posterior} guarantees; i.e.,   
the mathematical bounds are available before the learning is done, which provides insights a priori to understand the phenomenon and to design algorithms, but the numerical value of the bounds depend on the posterior quantities. A firm understanding of analytical learning theory requires a different style of thinking and a shift of technical basis from statistics (e.g., concentration inequalities) to measure theory.  We present the foundation of analytical learning theory in Section \ref{sec:basis} and  several applications in Sections \ref{sec:classical_ML}-\ref{sec:method}.

\section{Preliminaries}

 In machine learning, a typical goal is to return  a model  $\hat y_{\mathcal A(S_{m})}$ via a learning algorithm  $\mathcal A$ given a dataset $S_{m}=\{s^{(1)},\dots,s^{(m)}\}$ such that  the expected error   $\EE_{\mu}[L \hat y_{\mathcal A(S_{m})}]\triangleq \EE_{z}[L \hat y_{\mathcal A(S_{m})}(z)]$ with respect to  a true (unknown)  normalized measure $\mu$  is minimized. Here,    $L\hat y$ is a function that combines a loss function $\ell$ and a  model $\hat y$; e.g.,
in supervised learning,  $L\hat y(z)=\ell(\hat y(x),y)$, where $z=(x,y)$ is a pair of an input $x$ and a target $y$.
 Because the expected error $\EE_\mu[L  \hat y_{\mathcal A(S_{m})} ]$ is often not computable, we usually approximate the expected error by an empirical error  $\hat \EE_{Z_{m'}}[L \hat y_{\mathcal A(S_{m})}]\triangleq\frac{1}{m'}\sum_{i=1}^{m'} L \hat y_{\mathcal A(S_{m})}(z^{(i)})$ with a dataset $Z_{m'}=\{z^{(1)},\dots,z^{(m')}\}$. Accordingly, we define \textit{the generalization gap}
$\triangleq \EE_\mu[L  \hat y_{\mathcal A(S_{m})}] -  \hat \EE_{S_{m}}[L  \hat y_{\mathcal A(S_{m})}]$.  One of  the goals of learning theory is to explain and validate  when and how minimizing $\hat \EE_{S_{m}}[L \hat y_{\mathcal A(S_{m})}]$ is a sensible approach to minimizing $\EE_\mu[L  \hat y_{\mathcal A(S_{m})}]$ by analyzing the generalization gap, and to provide bounds on the performance of $\hat y_{\mathcal A(S_{m})}$ on new data.

\subsection{Discrepancy and variation}

In the following, we define a quality of a dataset, called \textit{discrepancy}, and a quality of a function, called \textit{variation in the sense of Hardy and Krause}. These definitions have been used in harmonic analysis, number theory, and numerical analysis  \citep{krause1903fouriersche,hardy1906double,hlawka1961funktionen,niederreiter1978quasi,aistleitner2017functions}. This study adopts these definitions  in the context of machine learning. Intuitively, the \textit{star-discrepancy} $D^{*}[T_{m},\nu]$ evaluates how well a  dataset $T_m=\{t^{(1)},\dots,t^{(m)}\}$ captures a normalized  measure $\nu$, and the \textit{variation $V[f]$ in the sense of Hardy and Krause}   computes how a function $f$ varies in total w.r.t.  each small perturbation of every cross combination of its variables.  

\subsubsection{Discrepancy of dataset with respect to a measure}   

 For any $t=(t_{1},\dots,t_{d})\in[0,1]^{d}$, let $B_t \triangleq [0,t_1] \times  \cdots  \times [0,t_d]$ be a closed axis-parallel box with one vertex at the origin. The \textit{local discrepancy} $D[B_t;T_{m},\nu]$ of a dataset $T_m=\{t^{(1)},\dots,t^{(m)}\}$ with respect to a normalized Borel measure $\nu$ on a set $B_t$ is defined as
\begin{displaymath}
D[B_t;T_m,\nu] \triangleq \left(\frac{1}{m}\sum_{i=1}^m \one_{B_{t}}(t^{(i)}) \right) -\nu \big(B_{t} \big) 
\end{displaymath}
where $\one_{B_{t}}$ is the indicator function of a set $B_{t}$. Figure \ref{fig:local_discrepancy} in Appendix \ref{app:illustration_discrepancy} shows an illustration of the  \textit{local discrepancy} $D[B_t;T_{m},\nu]$ and related notation. 
The \textit{star-discrepancy} $D^{*}[T_{m},\nu]$ of a dataset $T_m=\{t^{(1)},\dots,t^{(m)}\}$ with respect to a normalized Borel measure $\nu$ is defined as 
\begin{displaymath}
D^*[T_{m},\nu] \triangleq \sup_{t\in[0,1]^{d}} \big| D[B_t;T_m,\nu] \big|.
\end{displaymath}

\subsubsection{Variations of a function}
 Let $\partial_l$ be the partial derivative operator; that is, $
\partial_l g(t_{1},\dots,t_{k})
$
is the partial derivative of a function $g$ with respect to the $l$-th coordinate at a point $(t_{1},\dots,t_{k})$. Let  $\partial_{1,\dots,k}^k \triangleq\partial_{1},\dots,\partial_{k}$. A partition $P$ of $[0,1]^k$  with size $m_1^{P},\dots, m_k^P$ is a set of finite sequences $t^{(0)}_l,t^{(1)}_l\dots,t^{( m_l^P)}_l$ ($l=1,\dots,k$) such that $0=t^{(0)}_l \le t^{(1)}_l \le \cdots \le t^{(m_l^{P})}_l=1$ for $l=1,\dots,k$. We define a difference operator $\Delta^P_{l}$   with respect to a partition $P$ as: given a function $g$ and a point $(t_{1}, \dots t_{l-1},t_{l}^{(i)},t_{l+1},\dots, t_{k})$ in the partition $P$ (for $i=0,\dots,m_l^P-1$), 
\small
\begin{align*}
\Delta^P_{l} g(t_{1}, \dots t_{l-1},t_{l}^{(i)},t_{l+1},\dots, t_{k})  
 = g(t_{1}, \dots t_{l-1},t_{l}^{(i+1)},t_{l+1},\dots, t_{k})
 - g(t_{1}, \dots t_{l-1},t_{l}^{(i)},t_{l+1},\dots, t_{k}), 
\end{align*}
\normalsize
where $(t_{1}, \dots t_{l-1},t_{l}^{(i+1)},t_{l+1},\dots, t_{k})$ is the subsequent  point in the partition $P$   along the coordinate $l$. Let  $\Delta^P_{1,\dots,k}\triangleq\Delta^{P}_{1} \dots  \Delta^{P}_{k}$. 
Given a  function $f$ of $d$ variables, let $f_{j_{1}\dots j_k}$ be the  function restricted on $k\le d$ variables such that $f_{j_{1}\dots j_k}(t_{j_{1}},\dots,t_{j_k}) = f(t_{1},\dots,t_d)$, where  $t_{l} \equiv 1$ for all $l \notin \{j_{1},j_2,\dots j_k\}$. That is, $f_{j_{1}\dots j_k}$ is a function of $(t_{j_1}, \dots,t_{j_k})$ with other original variables being fixed to be one. 

The \textit{variation of $f_{j_{1}\dots j_k}$ on $[0,1]^k$ in the sense of Vitali} is defined as 
\begin{align*}
V^{(k)}[f_{j_{1}\dots j_k}]
\triangleq\ \sup_{P \in \mathcal{P}_k} \sum_{i_1=1}^{m^P_1-1} \dots \sum_{i_{k}=1}^{m^P_k-1} \left|\Delta^P_{1,\dots,k} f_{j_{1}\dots j_k}(t_{j_1}^{(i_{1})}, \dots,t_{j_k}^{(i_{k})}) \right|, 
\end{align*}
where $\mathcal{P}_k$ is the set of all partitions of $[0,1]^k$. The \textit{variation of $f$ on $[0,1]^d$ in the sense of Hardy and Krause} is defined as 
\begin{displaymath}
V[f] = \sum_{k=1}^d  \ \ \sum_{1\le j_1 <  \dots < j_k \le d} V^{(k)}[f_{j_{1}\dots j_k}]. 
\end{displaymath}

For example, if $f$ is linear on its domain, $V[f] = \sum_{1\le j_1 \le d} V^{(1)}[f_{j_{1}}]$ because $V^{(k)}[f_{j_{1}\dots j_k}]=0$ for all $k>1$. The following proposition might be helpful in intuitively understanding the concept of the variation as well as   in computing it when applicable.  All the proofs in this paper are presented in Appendix \ref{app:proofs}.

\begin{proposition} \label{prop:variation_diff}
Suppose that $f_{j_{1}\dots j_k}$ is a function for which $\partial_{1,\dots,k}^k f_{j_{1}\dots j_k}$ exists on $[0,1]^k$. Then, 
\small
$$
V^{(k)}[f_{j_{1}\dots j_k}] \le \sup_{(t_{j_{1}},\dots,t_{j_k})\in[0,1]^{k}} \left|\partial_{1,\dots,k}^k f_{j_{1}\dots j_k}(t_{j_{1}},\dots,t_{j_k})  \right|.
$$
\normalsize
If $\partial_{1,\dots,k}^k f_{j_{1}\dots j_k}$ is also continuous on $[0,1]^{k}$,
\small
$$
V^{(k)}[f_{j_{1}\dots j_k}] =  \int_{[0,1]^k} \left| \partial_{1,...,k}^k f_{j_{1}... j_k}(t_{j_{1}},...,t_{j_k})  \right| dt_{j_{1}} \hspace{-1pt} \cdot \cdot \cdot dt_{j_k}.
$$
\normalsize
\end{proposition}

\section{A basis of  analytical learning theory} \label{sec:basis}

This study considers the problem of analyzing the generalization gap $\EE_\mu[L \hat y_{\mathcal A(S_{m})}] -  \hat \EE_{S_{m}} \allowbreak [ L \hat y_{\mathcal A(S_{m})}]$ between the expected error $\EE_\mu[L \hat y_{\mathcal A(S_{m})}]$ and the training error $\hat \EE_{S_{m}}[L  \hat y_{\mathcal A(S_{m})}]$. For the purpose of   general applicability, our base theory analyzes a more general quantity, which is the generalization gap $\EE_\mu[L \hat y_{\mathcal A(S_{m})}] -  \hat \EE_{Z_{m'}}[L  \hat y_{\mathcal A(S_{m})}]$ between the expected error $\EE_\mu[L \hat y_{\mathcal A(S_{m})}]$ and any empirical error $\hat \EE_{Z_{m'}}[L  \hat y_{\mathcal A(S_{m})}]$  with any dataset $Z_{m'}$ (of size $m'$) including  the training dataset with  $Z_{m'}=S_{m}$.
 Whenever we write  $Z_{m'}$, it is always including the case of $Z_{m'}=S_{m}$; i.e.,  the  case where the model is evaluated on the training set.  

With our notation, one can observe  that the generalization gap is fully and deterministically specified by  a \textit{problem instance} $(\mu,S_{m},Z_{m'},L  \hat y_{\mathcal A(S_{m})})$, where we identify an omitted measure space  $(\Zcal,\Sigma,\mu)$ by the measure $\mu$ for brevity.  Indeed,  the expected error is defined by the Lebesgue integral of a function $L  \hat y_{\mathcal A(S_{m})}$ on a (unknown) normalized measure space  $(\Zcal,\Sigma,\mu)$  as 
$\EE_\mu[L  \hat y_{\mathcal A(S_{m})}] \allowbreak = \int_{\Zcal} L  \hat y_{\mathcal A(S_{m})} d\mu$, which is a deterministic mathematical object.
Accordingly, we introduce the following notion of  \textit{strong instance-dependence}: a mathematical object $\varphi$ is said to be  \textit{strongly instance-dependent} in  the theory of the generalization gap of the tuple $(\mu,S_{m},Z_{m'} \allowbreak, L  \hat y_{\mathcal A(S_{m})})$ if the object $\varphi$ is invariant under any  change of any mathematical  object that contains or depends on any $\bar \mu\neq \mu$, any $\hat y \neq  \hat y_{\mathcal A(S_{m})}$,  or any $\bar S_m$ such that $\bar S_{m}\neq S_{m}$ and $\bar S_{m}\neq Z_{m'}$.
Analytical learning theory is designed to provide  mathematical bounds and equations that are strongly instance-dependent. \vspace{2pt}

\subsection{Analytical decomposition of expected error} \label{sec:main_analytical_decomp}

Let  $(\Zcal,\Sigma,\mu)$ be any  (unknown) normalized measure space that defines the expected error, $\EE_{\mu}[L  \hat y]=\int_{\Zcal} L  \hat y \ d\mu$. Here, the measure space may correspond to an input-target pair as $\Zcal=\Xcal \times \Ycal$ for supervised learning,   the generative hidden space $\Zcal$ of $\Xcal \times \Ycal$ for unsupervised / generative models, or  anything else  of interest (e.g., $\Zcal=\Xcal$). Let $\mathcal T_* \mu$ be the pushforward measure of $\mu$ under a map $\mathcal T$. Let $\mathcal T(Z_{m'})=\{\mathcal T(z^{(1)}),\dots,\mathcal T(z^{(m')})\}$ be the image of the dataset $Z_{m'}$ under   $\mathcal T$. Let $|\nu|(E)$ be the total variation of a measure $\nu$ on  $E$. For vectors $a,b \in [0,1]^d$, let $[a,b]=\{t\in[0,1]^d: a \le t\le b\}$, where $\le$ denotes the product order; that is, $a \le t$ if and only if $a_j \le t_j$ for $j=1,\dots,d$. This paper adopts the convention that the infimum of the empty set is positive infinity.

Theorem \ref{thm:main} is introduced below to exploit the various structures in machine learning  through the decomposition $L  \hat y_{\mathcal A(S_{m})}(z) = (f \circ \mathcal T) (z)$ where $\mathcal T(z)$ is the output of a  representation function and $f$ outputs the associated loss. Here, $\mathcal T(z)$ can be any intermediate representation on the path from the raw
data  (when $\mathcal T(z)=z$) to the output  (when $\mathcal T(z) = L \hat y(z)$). The proposed theory holds true even if the representation $\mathcal T(z)$ is learned. The empirical error $\hat \EE_{Z_{m'}}[L  \hat y_{\mathcal A(S_{m})}] $ can be the training error with $Z_{m'}=S_{m}$ or the test/validation error with $Z_{m'}\neq S_{m}$.

\begin{theorem} \label{thm:main}
For  any $L  \hat y$, let $\mathcal{F}[L  \hat y]$ be a set of all  pairs  $(\mathcal T,f)$ such that  $\mathcal T:(\Zcal, \Sigma) \rightarrow ([0,1]^{d}, \mathcal B([0,1]^d))$ is a measurable function,  $f:([0,1]^{d}, \mathcal B([0,1]^d))\rightarrow (\RR, \mathcal{B}(\RR))$  is of   bounded variation as $V[f] < \infty$, and
$$
L  \hat y(z)= (f \circ \mathcal T) (z) \ \ \  \text{ for all } z \in \Zcal,
$$
where $\mathcal{B}(A)$ indicates the Borel $\sigma$-algebra on $A$.
Then, for any dataset pair $(S_{m},Z_{m'})$ (including $Z_{m'}=S_{m}$) and any $L  \hat y_{\mathcal A(S_{m})}$, 
\begin{enumerate}[label=(\roman*)]
\item 
$
\!
\begin{aligned}[t]
\EE_{\mu}[L  \hat y_{\mathcal A(S_{m})}] 
\le \hat \EE_{Z_{m'}}[L  \hat y_{\mathcal A(S_{m})}] + \hspace{-3pt}  \inf_{\substack{(\mathcal T,f) \in\hat {\mathcal F}}} \hspace{-2pt} V[f] \; D^{*}[\mathcal T_{*}\mu, \mathcal T(Z_{m'})], 
\end{aligned} 
$

where $\hat {\mathcal F}=\mathcal{F}[L  \hat y_{\mathcal A(S_{m})}]$, and
\item 
for any $(\mathcal T,f) \in\mathcal{F}[L  \hat y_{\mathcal A(S_{m})}]$ such that  $f$ is  right-continuous component-wise,
\small 
\begin{align*}
\EE_{\mu}[L  \hat y_{\mathcal A(S_{m})}] = \hat \EE_{Z_{m'}}[L  \hat y_{\mathcal A(S_{m})}] 
+ \int_{[0,1]^d} \left( (\mathcal T_{*}\mu)([\bold 0,t])-\frac{1}{m'}\sum_{i=1}^{m'}\one_{[\bold 0,t]}(\mathcal T(z_{i})) \right) d\nu_{f}(t), 
\end{align*}
\normalsize
where $z_i \in Z_{m'}$, and $\nu_{f}$ is a signed measure corresponding to $f$ as $f(t)= \nu_{f}([t,\bold 1])+f(\bold 1)$ and $| \nu_{f}|([0,1]^d)=V[f]$.   
\end{enumerate} 

\end{theorem}

The statements in Theorem \ref{thm:main} hold for each individual instance $(\mu,S_{m},Z_{m'},L  \hat y_{\mathcal A(S_{m})})$, for example, without taking a supremum over a set of other instances.  In contrast, typically in previous bounds, when asserting that  an upper bound holds on $\EE_{\mu}[L  \hat y] - \hat \EE_{S_{m}}[L  \hat y]$  for any $\hat y \in \mathcal H$ (with high probability), what it  means is  that the upper bound holds on $\sup_{\hat y \in \mathcal H}(\EE_{\mu}[L  \hat y] - \hat \EE_{S_{m}}[L  \hat y])$ (with high probability). Thus, in classical bounds including   data-dependent ones,  as   $\mathcal H$ gets larger and more complex, the bounds tend to become more pessimistic for the actual instance  $\hat y_{\mathcal A(S_{m})}$ (learned with the actual instance  $S_{m}$), which is avoided in Theorem \ref{thm:main}.

\begin{remark} \label{rem:imp_subtle}
The bound  and the equation in Theorem \ref{thm:main} are strongly instance-dependent, and in particular, \textit{invariant to  hypothesis space $\mathcal H$ and the properties of learning algorithm $\mathcal A$ over datasets different from a given training dataset $S_{m}$} (and $Z_{m'}$). 
\end{remark}

\begin{remark} \label{rem:consequence_of_remark_imp_subtle}
Theorem \ref{thm:main} together with Remark \ref{rem:imp_subtle}  has an immediate practical consequence. For example, even if the true model is  contained in some  ``small'' hypothesis space $\mathcal H_1$, we might want to use a much more complex ``larger'' hypothesis space $\mathcal H_2$ in practice such that the optimization becomes easier and the training trajectory reaches a better model  $\hat y_{\mathcal A(S_{m})}$ at the end of the learning process (e.g., over-parameterization in deep learning potentially makes the non-convex optimization easier;  see \citealt{dauphin2014identifying,choromanska2015loss,soudry2017exponentially}). This is consistent with   both Theorem \ref{thm:main} and practical observations in  deep learning, although it can be puzzling from the viewpoint of previous   results  that explicitly or implicitly penalize the use of more complex ``larger'' hypothesis   spaces (e.g., see \citealt{zhang2016understanding}).
\end{remark}

\begin{remark} \label{rem:comparion_infeasible}
Theorem \ref{thm:main} does not require statistical assumptions. Thus, it is applicable even when  statistical assumptions required by statistical learning theory are violated in practice.     
\end{remark}

Theorem \ref{thm:main} produces bounds that can be zero even with  $m=1$ (and $m'=1$) (as an examples are provided  throughout the paper),  supporting the concept of one-shot learning. This is true, even if the dataset is not drawn according to the measure $\mu$. This is because although  such a dataset may incur a  lager value of  $D^*$ (than a usual i.i.d. drawn dataset), it can decrease $V[f]$ in the generalization bounds of $V[f] D^{*}[\mathcal T_{*}\mu, \mathcal T(S_{m})]$. 
Furthermore, by being strongly instance-dependent on the learned model $  \hat y_{\mathcal A(S_{m})}$, Theorem \ref{thm:main} supports the concept of curriculum learning~\citep{Bengio-et-al-ICML2009}. This is because   curriculum learning directly guides the learning to obtain a good model $\hat y_{\mathcal A(S_{m})}$, which minimizes $V[f]$ by its definition.

\subsection{Additionally using statistical assumption and general bounds on $D^*$} \label{sec:general-bound_star-discrepancy}

By additionally using the standard i.i.d. assumption, Proposition \ref{prop:random_sample}  provides a general bound on the star-discrepancy $D^*[\mathcal T_{*}\mu, \mathcal T(Z_{m'})]$ that appears in Theorem \ref{thm:main}.
It is a direct consequence of \citep[Theorem 2]{heinrich2001inverse}.

\begin{proposition} \label{prop:random_sample}
Let \small $\mathcal T(Z_{m'})= \{\mathcal T(z^{(1)}),\dots,\mathcal T(z^{(m')})\}=\{t^{(1)},\dots,t^{(m')}\}$ \normalsize be a set of i.i.d. random variables with values on $[0,1]^{d}$ and distribution $\mathcal T_{*}\mu$. Then, there exists a positive constant $c_{1}$ such that for all $m'\in\mathbb{N^{+}}$ and all $c_{2} \ge c_{1}$, with probability at least $1-\delta$, 
$$
D^*[\mathcal T_{*}\mu, \mathcal T(Z_{m'})] \le c_{2} \sqrt{\frac{d}{m'}} $$
where $\delta=\frac{1}{c_{2} \sqrt d} (c_{1}c_{2}^{2}e^{-2c_{2}^2})^d$ with $c_{1}c_{2}^{2}e^{-2c_{2}^2}<1$. 
\end{proposition}
\begin{remark} Proposition \ref{prop:random_sample} is not probabilistically vacuous in the sense that we can  increase $c_2$ to obtain $1-\delta>0$, at the cost of increasing the constant $c_2$ in the bound. Forcing $1-\delta>0$ still keeps $c_2$ constant without dependence on relevant variables such as $d$ and $m'$. This is  because $1-\delta>0$ if $c_2$ is large enough such that $c_{1}c_{2}^{2} <e^{2c_{2}^2}$, which  depends only on the constants.
\end{remark}
Using Proposition \ref{prop:random_sample}, one can immediately provide a statistical bound via Theorem \ref{thm:main} over random  $Z_{m'}$. To see how  such a result differs from that of statistical learning theory, consider the case of $Z_{m'}=S_{m}$. That is, we are looking at classic training error. Whereas statistical learning theory applies a statistical assumption to the whole object  $
\EE_{\mu}[L  \hat y_{\mathcal A(S_{m})}]  -\hat \EE_{S_{m}}[L  \hat y_{\mathcal A(S_{m})}]$, analytical learning theory first decomposes $\EE_\mu[L  \hat y_{\mathcal A(S_{m})}]  -\hat \EE_{S_{m}}[L  \hat y_{\mathcal A(S_{m})}]$ into $V[f]D^*[\mathcal T_{*}\mu, \mathcal T(S_{m})]$ and then applies the statistical assumption only to $D^*[\mathcal T_{*}\mu, \mathcal T(S_{m})]$. \textit{This makes $V[f]$   strongly instance-dependent  even with the
statistical assumption.} For example, with $f(z)=L  \hat y_{\mathcal A(S_{m})}(z)$ and $\mathcal T(z)=z$, if the training dataset $S_{m}$  satisfies the standard i.i.d. assumption, we have that with high probability, 
\begin{align} \label{eq:stat_assumption_0}
\EE_\mu[L  \hat y_{\mathcal A(S_{m})}] -  \hat \EE_{S_{m}}[L  \hat y_{\mathcal A(S_{m})}] \le c_{2} V[L  \hat y_{\mathcal A(S_{m})}]  \sqrt{\frac{d}{m}}, 
\end{align}
where the term $V[L  \hat y_{\mathcal A(S_{m})}]$ is  strongly instance-dependent.  

In Equation \eqref{eq:stat_assumption_0}, it is unnecessary for  $m$ to approach infinity in order  for the generalization gap to go to zero. As an extreme example, if the variation of $\hat y_{\mathcal A(S_{m})}$ aligns with that of the true $y$ (i.e., $L  \hat y_{\mathcal A(S_{m})}$ is constant), we have that $V[L  \hat y_{\mathcal A(S_{m})}]=0$ and the generalization gap becomes zero even with $m=1$. This example illustrates the fact that Theorem \ref{thm:main}  supports the concept of one-shot learning via the transfer of knowledge into the resulting model $\hat y_{\mathcal A(S_{m})}$.

For the purpose of the non-statistical decomposition of $\EE_\mu[L  \hat y_{\mathcal A(S_{m})}]  -\hat \EE_{S_{m}}[L  \hat y_{\mathcal A(S_{m})}]$,  instead of Theorem \ref{thm:main}, we might  be tempted to  conduct a   simpler decomposition with the H\"{o}lder inequality  or its variants. However, such a simpler decomposition is dominated by a  difference between the true  measure and the empirical measure on an arbitrary set  in high-dimensional space, which suffers from the curse of dimensionality. Indeed, the proof of Theorem \ref{thm:main} is devoted to reformulating $\EE_\mu[L  \hat y_{\mathcal A(S_{m})}]  -\hat \EE_{S_{m}}[L  \hat y_{\mathcal A(S_{m})}]$ via the equivalence in the measure and the variation \textit{before taking any inequality}, so that we can avoid such an issue. That is, the star-discrepancy evaluates the difference in the measures on high-dimensional boxes with one vertex at the origin, instead of on an arbitrary set.

The following proposition proves the existence of a  dataset $Z_{m'}$  with a   convergence rate of $D^*[\mathcal T_{*}\mu, \mathcal T(S_{m})]$ that  is asymptotically faster than $O(\sqrt{1/m})$ in terms of the dataset size $m'$. This is a direct consequence of \citep[Theorem 2]{aistleitner2014low}.

\begin{proposition} \label{prop:determ_sample}
Assume that $\mathcal T$ is a surjection. Let $\mathcal T_{*}\mu$ be any  (non-negative) normalized Borel measure on $[0,1]^d$. Then, for any $m' \in \NN^+$, there exists a dataset  $Z_{m'}$ such that 
$$
D^*[\mathcal T_{*}\mu, \mathcal T(Z_{m'})] \le 63 \sqrt d \frac{(2+\log_2 m')^{(3d+1)/2}}{m'}.
$$   
\end{proposition}

This can be of  interest when we can choose $\mathcal T$ to make $d$ small without increasing $V[f]$ too much; i.e.,  it then provides a faster convergence rate than usual statistical guarantees. If $\Zcal \subseteq \mathbb{R}^{d_z}$ (which is true in many practical cases),   we can have $d=1$ by setting $\mathcal T: \Zcal \rightarrow [0,1]$, because there exists a bijection between the interior of $\Zcal $ and $(0,1)$. Then, although the variation of $\mathcal T$ is unbounded in general, $V[f]$  might be still  small. For example, it is still zero if the variation of $\hat y_{\mathcal A(S_{m})}$ aligns with that of the true $y$ in this space of $[0,1]$.

\subsection{General examples}

The following example  provides insights on the quality of \textit{learned} representations:

\begin{example} \label{example:hidden}
 
Let $\mathcal T(z)=(\phi(x), v)$ where $\phi$ is a map of any \textit{learned} representation and $v$ is a variable such that  there exists a function $f$ satisfying $L\hat y_{\mathcal A(S_{m})}(z)=f(\phi(x), v)$  (for supervised learning, setting $v:=y$ always satisfies this condition regardless of the information contained in $\phi(x)$). For example, $\mathcal \phi(x)$ may represent the output of  any intermediate hidden layer in deep learning (possibly the last hidden layer), and $v$ may encode the noise left in the label $y$. Let $f$ be a map such that $L\hat y_{\mathcal A(S_{m})} = f(\mathcal T(z))$.  Then, if $V[f]<\infty$, Theorem \ref{thm:main} implies that for any dataset pair $(S_{m},Z_{m'})$ (including $Z_{m'}=S_{m}$),
$\EE_{\mu}[L\hat y_{\mathcal A(S_{m})}] \le \hat \EE_{Z_{m'}}[L\hat y_{\mathcal A(S_{m})}] +V[f] D^*[\mathcal T_*\mu,\mathcal T(Z_{m'})].
$\end{example}

Example \ref{example:hidden}  partially supports the concept of the \textit{disentanglement} in deep learning \citep{bengio2009learning} and proposes a new concrete method to measure the  degree  of  disentanglement as follows. In the definition of \small $V[f] = \sum_{k=1}^d  \ \ \sum_{1\le j_1 <  \dots < j_k \le d} V^{(k)}[f_{j_{1}\dots j_k}]$\normalsize, each term $V^{(k)}[f_{j_{1}\dots j_k}]$  can be viewed as measuring  how \textit{entangled} the $j_{1},\dots ,j_k$-th variables are in a space of a learned (hidden) representation. We can observe  this from the definition of $V^{(k)}[f_{j_{1}\dots j_k}]$ or from Proposition \ref{prop:variation_diff} as: \small $V^{(k)}[f_{j_{1}\dots j_k}] =  \int_{[0,1]^k} \left| \partial_{1,\dots,k}^k f_{j_{1}\dots j_k}(t_{j_{1}},\dots,t_{j_k})  \right| dt_{j_{1}} \cdots dt_{j_k}$\normalsize, where \small $\partial_{1,\dots,k}^k  f_{j_{1}\dots j_k}(t_{j_{1}},\dots,t_{j_k})$ \normalsize is the $k$-th order \textit{cross} partial derivatives across the $j_{1},\dots ,j_k$-th variables. 
If all the variables in a space of a  learned  (hidden) representation are completely disentangled in this sense, $V^{(k)}[f_{j_{1}\dots j_k}]=0$ for all $k\ge 2$ and $V[f]$ is minimized to $V[f]=\sum_{j_1=1}^dV^{(1)}[f_{j_{1}}]$.  Additionally, Appendices \ref{app:patho_non-flat} and \ref{app:high_deriv} provide discussion of the effect of flatness in measures and higher-order derivatives.

One of the reasons why analytical learning theory is complementary to statistical learning theory is the fact that we can  naturally combine the both. For example, in Example \ref{example:hidden}, we cannot directly adopt the probabilistic bound on $ D^{*}[\mathcal T_{*}\mu, \mathcal T(S_{m})]$ from Section \ref{sec:general-bound_star-discrepancy}, if  $\mathcal T(S_{m})$ does  not satisfy the i.i.d. assumption because $\mathcal T$ depends on the whole dataset $S_{m}$. In this case, to analyze $D^{*}[\mathcal T_{*}\mu, \mathcal T(S_{m})]$, we can  use the approaches in statistical learning theory, such as Rademacher complexity or covering number. 
 To see this, consider a set $\Phi$    such that $\mathcal T \in \Phi$ and $\Phi$ is independent of $S_{m}$. Then, by applying Proposition \ref{prop:random_sample} with a union bound over a cover  of $\Phi$, we can obtain probabilistic bounds on $D^*$ with the log of  the \textit{covering number }of $\Phi$ for all representations $\mathcal T' \in \Phi$. As in data-dependent approaches (e.g., \citealt[Lemma A.9]{bartlett2017spectrally}), one can also consider a sequence of sets $\{\Phi_j\}_j$ such that $\mathcal T \in \cup_j\Phi_j$, and one can obtain a data-dependent bound on $D^{*}[\mathcal T_{*}\mu, \mathcal T(S_{m})]$ via a  complexity of $\Phi_j$.

The following example establishes the  tightness of Theorem \ref{thm:main} (i) with the 0-1 loss in general, where $\iota:\{0,1\} \rightarrow [0,1]$ is an inclusion map:     
\begin{example} \label{example:loss}
 Theorem \ref{thm:main} (i) is tight in   multi-class classification with  0-1 loss as follows. Let  $\mathcal T= \iota \circ L{\hat y_{\mathcal A(S_{m})}}$. Let $f$ be an identity map. 
 Then,  $V[f]=1$ and $L{\hat y_{\mathcal A(S_{m})}}(z)= (f \circ \mathcal T) (z)$ for all $z \in \Zcal$.
Then, the pair of $\mathcal T$ and $f$ satisfies the condition in Theorem \ref{thm:main} as $L{\hat y_{\mathcal A(S_{m})}}$ and $\iota$ are measurable functions. Thus, from Theorem \ref{thm:main}, 
$\EE_{\mu}[L  \hat y_{\mathcal A(S_{m})}] -  \hat \EE_{Z_{m'}}[L  \hat y_{\mathcal A(S_{m})}] 
\le V[f] D^*[\mathcal T_{*}\mu, \mathcal T(Z_{m'})]
= |(\mathcal T_{*}\mu)(\{1\})-\EE_{Z_{m'}}[L  \hat y_{\mathcal A(S_{m})}]|)
$ (see Appendix \ref{app:proof_gexample-loss} for this derivation),
which establishes a tightness of Theorem \ref{thm:main} (i) with the 0-1 loss as follows: for any dataset pair $(S_{m},Z_{m'})$ (including $Z_{m'}=S_{m}$),
$
\left| 
\EE_{\mu}[L  \hat y_{\mathcal A(S_{m})}] -  \hat \EE_{Z_{m'}}[L  \hat y_{\mathcal A(S_{m})}] \right| =V[f] D^*[\mathcal T_{*}\mu, \mathcal T(Z_{m'})].
$
\end{example}    

The following example applies Theorem \ref{thm:main} to a raw representation space $\mathcal T(z)=z$ and a loss space  $\mathcal T(z)=(\iota \circ L{\hat y_{\mathcal A(S_{m})}})(z)$:
\begin{example}  \label{example:1}
Consider a normalized domain $\Zcal  =  [0,1]^{d_z}$ and a Borel measure $\mu$ on $\Zcal$. For example, $\Zcal$ can be an unknown hidden generative space or an input-output space ($\Zcal=\Xcal \times \Ycal$). Let us  apply  Theorem \ref{thm:main} to this measure space with $\mathcal T(z)=z$ and $f=L\hat y_{\mathcal A(S_{m})}$. Then, if $V[L\hat y_{\mathcal A(S_{m})}]<\infty$, Theorem \ref{thm:main} implies that
for any dataset pair $(S_{m},Z_{m'})$ (including $Z_{m'}=S_{m}$) and  any $L  \hat y_{\mathcal A(S_{m})}$, 
$\EE_{\mu}[L\hat y_{\mathcal A(S_{m})}] \le \hat \EE_{Z_{m'}}[L\hat y_{\mathcal A(S_{m})}] +V[L\hat y_{\mathcal A(S_{m})}] D^*[\mu,Z_{m'}]$.  
\end{example} 

Example \ref{example:1}  indicates  that we can regularize $V[L\hat y_{\mathcal A(S_{m})}]$ in some space  $\Zcal$ to control the generalization gap. For example, letting the model $\hat y_{\mathcal A(S_{m})}$ be invariant to a subspace that is not essential for prediction decreases the bound on $V[L\hat y_{\mathcal A(S_{m})}]$. As an extreme example, if  $x=g(y,\xi)$ with some  generative function $g$ and noise $\xi$ (i.e., a setting considered in an information theoretic approach),  $\hat y_{\mathcal A(S_{m})}$ being invariant to $\xi$ results in a smaller bound on $V[L\hat y_{\mathcal A(S_{m})}]$. This is qualitatively related to an information theoretic observation such as in \citep{achille2017emergence}.

\section{Application to linear regression} \label{sec:classical_ML}

Even in the classical setting of linear regression, recent papers (\citealt{zhang2016understanding}, Section 5; \citealt{kawaguchi2017generalization}, Section 3; \citealt{poggio2017theory}, Section 5) suggest the  need for further theoretical studies to better understand  the question of precisely   what makes a learned model generalize well, especially with  an \textit{arbitrarily} rich hypothesis space and algorithmic instability. Theorem \ref{thm:main}  studies the question abstractly for machine learning in general. As a simple concrete example, this section considers linear regression. However, note that the theoretical results  in this section can be directly applied to deep learning as described in Remark \ref{rem:linear_to_dl}.

Let $S_{m}=\{s^{(i)}\}_{i=1}^m$ be a training dataset of the input-target pairs where $s^{(i)}=(x^{(i)},y^{(i)})$. Let   $\hat y_{\mathcal A(S_{m})}=\hat W\phi(\cdot)$ be the learned model at the end of any training process. For example, in empirical risk minimization, the matrix $\hat W$ is an output of the training process,  $\hat W :=\argmin_W \hat \EE_{S_{m}} [\frac{1}{2}\|W \allowbreak \phi(x)-y\|_2^2]$. Here, $\phi:(\mathcal X,\Sigma_x) \rightarrow ([0,1]^{d_\phi},\mathcal B([0,1]^{d_\phi}))$ is any normalized   measurable function, corresponding to fixed features. For any given variable $v$, let $d_v$ be the dimensionality of the variable $v$. The goal is to minimize the expected error $\EE_{s} [\frac{1}{2}\|\hat W\phi(x)-y\|_2^2]$  of the learned model $\hat W\phi(\cdot)$.

\subsection{Domains  with linear Gaussian labels} \label{sec:fixed_feature_true_model}

In this subsection only, we assume that the target output $y$ is structured  such that
$
y = W^* \phi(x) + \xi,
$
where $\xi$ is a zero-mean random variable independent of $x$. Many columns of $W^*$ can be zeros (i.e., sparse) such that  $W^* \phi(x)$  uses a small portion of the feature vector $\phi(x)$. Thus, this  label assumption   can be satisfied by including a sufficient number of  elements from a basis with uniform approximation power (e.g., polynomial basis, Fourier basis, a set of step functions, etc.) to the feature vector $\phi(x)$ up to a desired approximation error. Note that we do not assume any knowledge of $W^*$.

Let $\mu_x$ be the (unknown) normalized measure for the input $x$ (corresponding to the marginal distribution of $(x,y)$). Let $X_m=\{x^{(i)}\}_{i=1}^m$ and $\tilde S_{m}=\{(x^{(i)},\xi^{(i)})\}_{i=1}^m$ be the input part and the (unknown) input-noise part of the same training dataset as $S_{m}$, respectively. We do not assume  access to $\tilde S_{m}$. Let $W_{l}$ be the $l$-th column of the matrix $W$.

\begin{theorem} \label{thm:fixed_feature_1}
Assume that the labels are structured as described above  and  $\|\hat W - W^*\|<\infty$. Then,  Theorem \ref{thm:main} implies that
\begin{align} \label{eq:fixed_feature_1}
\EE_s\left[\frac{1}{2}\|\hat W\phi (x)- y\|_2^2\right] - \hat \EE_{S_{m}} \left[\frac{1}{2}\|\hat W\phi (x)- y\|_2^2\right]
\le V[f]D^{*}[\phi_*\mu_x,\phi(X_{m})] + A_1 + A_2 , 
\end{align}
where $f(t)=\frac{1}{2}\|\hat Wt-W^* t\|_2^{2}$, $A_1 = \hat \EE_{\tilde S_{m}}[\xi^\top (\hat W-W^{*}) \phi(x)]$,  $A_2 =\EE_{\xi}[\|\xi\|^2_2]- \hat \EE_{\tilde S_{m}}[\|\xi\|^2_2] $, and 
\begin{align*} 
\nonumber V[f]  \le  \sum_{l=1}^{d_{\phi}} \|(\hat W_{l} - W_{l}^*)^\top (\hat W - W^*)\|_1
 + \sum_{1\le l < l' \le d_{\phi}} |(\hat W_{l} - W_{ l}^*)^\top (\hat W_{l'} - W_{l'}^*)|.
\end{align*}

\end{theorem}

\begin{remark} \label{rem:fixed_feature_1}
 Theorem \ref{thm:fixed_feature_1} is tight in terms  of both the minimizer and its value, which is explained below. The bound in Theorem \ref{thm:fixed_feature_1} (i.e., the right-hand-side of Equation \eqref{eq:fixed_feature_1}) is minimized (to be the noise term  $A_2$ only) if and only if $\hat W=W^*$ (see Appendix \ref{app:supp_rem_fixed_feature_1}   for pathological cases). Therefore, minimizing the bound in  Theorem \ref{thm:fixed_feature_1} is equivalent to minimizing  the expected error \small$\EE_s[\|\hat W\phi (x)- y\|_2^2]$ \normalsize or generalization error (see Appendix \ref{app:supp_rem_fixed_feature_1} for further  details). Furthermore, the bound in Theorem \ref{thm:fixed_feature_1} holds with equality if  $\hat W=W^*$. Thus, the bound is tight in terms  of the minimizer and its value. 
\end{remark}

\begin{remark} \label{rem:fixed_feature_1_noize-term} 
For $D^{*}[\phi_*\mu_x,\phi(X_{m})]$ and $A_2$, we can straightforwardly  apply the probabilistic bounds under the standard i.i.d. statistical assumption. From Proposition \ref{prop:random_sample},
with high probability, $
D^{*}[\phi_*\mu_x, \allowbreak\phi(X_{m})] \le O (\sqrt{d_{\phi}/m}). 
$
From Hoeffding's inequality with  $M \ge \|\xi\|^2_2$,  with probability at least $1-\delta$,  $A_2\le M\sqrt{\ln(1/\delta)/2m}$. 
\end{remark}

It is not necessary for  $D^{*}[\phi_*\mu_x,\phi(X_{m})]$  to approach  zero to minimize the expected error; irrespective of whether the  training dataset satisfies a certain  statistical assumption to bound $D^{*}[\phi_*\mu_x,\allowbreak\phi(X_{m})]$, we can minimize the expected error via making $\hat W$ closer to $W^*$ as shown in Theorem \ref{thm:fixed_feature_1}.

\subsection{Domains  with unstructured/random labels} \label{sec:fixed_feature_no_struc}

In this subsection, we discard the linear Gaussian label assumption in the previous subsection and consider the worst case scenario where
$y$ is a variable independent of  $x$. 
This corresponds to the random label experiment by \citet{zhang2016understanding}, which posed another  question: how to theoretically distinguish the generalization behaviors with  structured labels from those with  random labels. Generalization behaviors in practice are expected to be  significantly different in problems with structured labels or random labels, even when the hypothesis space and learning algorithm  remain unchanged.

As desired, Theorem \ref{thm:fixed_feature_2} (unstructured labels) predicts a completely different generalization behavior from that in Theorem \ref{thm:fixed_feature_1} (structured labels), even with an identical hypothesis space and learning algorithm. Here,
we consider the normalization of $y$ such that  $y \in [0,1]^{d_y}$. Let $\mu_s$ be the (unknown) normalized measure for the pair $s=(x,y)$.

\begin{theorem} \label{thm:fixed_feature_2}
\emph{}
Assume  unstructured labels as described above. Let $M=\sup_{t\in[0,1]}\|\hat Wt - y\|_{\infty}$. Assume that $\|\hat W\|<\infty$ and $M<\infty$. Then, Theorem \ref{thm:main} implies that
\begin{align} \label{eq:fixed_feature_2}
\EE_{s}\left[\frac{1}{2}\|\hat W\phi (x)- y\|_2^2\right] - \hat \EE_{S_{m}} \left[\frac{1}{2}\|\hat W\phi (x)- y\|_2^2\right]
 \le V[f]D^{*}[\mathcal T_*\mu_{s}, \mathcal T(S_{m})], 
\end{align}
where  $\mathcal T(s) = (\phi(x),y)$, $f(t,y)=\frac{1}{2}\|\hat Wt-y\|_2^2$, and
$$
V[f] \le (M+1) \sum_{l=1}^{d_\phi}\|\hat W_{l}\|_1 +\sum_{1\le l<l' \le d_\phi} |\hat W^\top_{l} \hat W_{l'}|+ d_{y} M.
$$
\end{theorem}

Unlike in the structured case (Theorem \ref{thm:fixed_feature_1}),  minimizing the bound on the generalization gap in the unstructured case requires us to control the norm of $\hat W$, which corresponds to the traditional results from statistical learning theory. As in statistical learning theory, the  generalization gap in Theorem \ref{thm:fixed_feature_2} (unstructured labels)  goes to zero as $D^{*}[\mathcal T_*\mu_{s}, \mathcal T(S_{m})]$ approaches zero via certain statistical assumption: e.g., via Proposition \ref{prop:random_sample}, with high probability, $D^{*}[\mathcal T_*\mu_{s}, \mathcal T(S_{m})]\le O (\sqrt{(d_{\phi}+d_y)/m})$.  This is in contrast to Theorem \ref{thm:fixed_feature_1} (the structured case) where we require no statistical assumption for the generalization gap to approach zero within polynomial sample complexity.

\begin{remark} \label{rem:linear_to_dl}
(Theorems  \ref{thm:fixed_feature_1} and \ref{thm:fixed_feature_2}  on representation learning)
\textit{Theorems  \ref{thm:fixed_feature_1} and \ref{thm:fixed_feature_2}  hold true,  even with learned representations $\phi$, instead of fixed features}. Let $\mathcal \phi(x)$ represent the last hidden layer in a neural network or the learned representation in representation learning in general. Consider the squared loss (square of output minus target). Then, the identical proofs of Theorems  \ref{thm:fixed_feature_1} and \ref{thm:fixed_feature_2} work with the learned representation $\phi$.
\end{remark}

\section{From analytical learning theory to methods in deep learning} \label{sec:method}
This section  further demonstrates the practical relevance of  analytical learning theory by showing its application to derive empirical methods. The complete code of our method and experiments is publicly available at \url{https://github.com/Learning-and-Intelligent-Systems/Analytical-Learning-Theory}.

\subsection{Theory}
We consider multi-class classification with a set $Y$ of class labels. Then,  
 
\begin{align*}
&\EE_{\mu}[L  \hat y_{\mathcal A(S_{m})}] - \hat \EE_{S_m}[L  \hat y_{\mathcal A(S_{m})}] 
\\ &= \sum_{y \in Y} p(y) \EE_{\mu_{x|y}} [L  \hat y_{\mathcal A(S_{m})}]-  \hat p(y) \hat \EE_{S_{x|y}}[L  \hat y_{\mathcal A(S_{m})}] \pm    p(y) \hat \EE_{S_{x|y}}[L  \hat y_{\mathcal A(S_{m})}]
\\ & = \sum_{y \in Y} p(y)\left( \EE_{\mu_{x|y}} [L  \hat y_{\mathcal A(S_{m})}]-    \hat \EE_{S_{x|y}}[L  \hat y_{\mathcal A(S_{m})}] \right) + \left(p(y)-\hat p(y) \right)  \hat \EE_{S_{x|y}}[L  \hat y_{\mathcal A(S_{m})}],
\end{align*}
where $\hat p(y) \triangleq \frac{|S_{x|y}|}{m}$, $ \hat \EE_{S_{x|y}}[L  \hat y_{\mathcal A(S_{m})}] \triangleq \frac{1}{|S_{x|y}|} \sum_{x \in S_{x|y}} L  \hat y_{\mathcal A(S_{m})}(x,y)$, and $S_{x|y} \subseteq S_m$ is the set of the training input points $x$ of the label $y$. Within the sum over $y$, by applying Theorem 1 (i) to each first term and Hoeffding's inequality to each second term, we have that with probability at least $1-\delta/2$,
\begin{align*}
&\EE_{\mu}[L  \hat y_{\mathcal A(S_{m})}] - \hat \EE_{S_m}[L  \hat y_{\mathcal A(S_{m})}] 
\\ &\le \sum_{y \in Y} p(y)\inf_{\substack{(\mathcal T_{y},f_{y}) \in\hat {\mathcal F_{y}}}} \hspace{-2pt} V[f_{y}] D^{*}[(\mathcal T_{y})_{*}\mu_{x|y}, \mathcal T_{y}(S_{x|y})] +  \hat \EE_{S_{x|y}}[L  \hat y_{\mathcal A(S_{m})}] \sqrt{\frac{\log 2/\delta}{2|S_{x|y}|}}.    
\end{align*}

Assume that there exists a generative (unknown) hidden space $G:(y,z) \mapsto x$ where the true label of the input $x=G(y,z)$ is  $y$ for any $z$ in the its normalized domain. We now set $\mathcal T_{y}:(x,y) \mapsto z$ and $f_y:z\mapsto L  \hat y_{\mathcal A(S_{m})}(x,y)$ where $z$ is the \textit{unknown}  hidden space that does not affect the true label. This choice does not depend on the dataset although it is unknown. Thus, by applying Proposition 2 with these $(T_{y},f_y)$, we have that with probability at least $1-\delta$,          
\begin{align} \label{eq:method}
\EE_{\mu}[L  \hat y_{\mathcal A(S_{m})}] - \hat \EE_{S_m}[L  \hat y_{\mathcal A(S_{m})}] \le\sum_{y \in Y}  c_{2}p(y)V[f_{y}] \sqrt{\frac{d_{z}}{|S_{x|y}|}}+  \hat \EE_{S_{x|y}}[L  \hat y_{\mathcal A(S_{m})}] \sqrt{\frac{\log 2/\delta}{2|S_{x|y}|}},
\end{align}
where $d_z$ is the dimensionality of the generative hidden space of $z$ and $c_2$ is a constant defined in  Proposition 2. 

Equation \ref{eq:method} tells us that if $V[f_{y}]$ is  bounded by a constant, the generalization error  \textit{goes to zero in polynomial sample complexity}  \textit{even with an arbitrarily complex hypothesis space and non-stable learning algorithm.} If the loss is 0-1 loss,   $V[f_y]=0$ when $(y,z)\mapsto   \hat y_{\mathcal A(S_{m})}(x)$ is invariant over $z$. In other words, to control $V[f_y]$, we want to have a model that is more invariant over the  space of $z$, which intuitively makes sense.        

\subsection{Methods}

The above result provides a theoretical basis for a family of \textit{consistency-based regularization} methods, including $\Pi$-Model ~\citep{laine}, virtual adversarial training ~\citep{takeru} and regularization with stochastic transformations and perturbations ~\citep{sajjadi}. These consistency-based regularization methods have been empirically   successful heuristics. These  methods are based on the intuition that perturbations of a data point $x \mapsto\ \tilde x $ should not change the output of a  model as $\hat y(x) \approx \hat y( \tilde x)$ if the true label is invariant under the perturbation; i.e., $y^*(x)=y^*(\tilde x)$ where $y^{*}$ outputs a correct label.
This intuitive goal is achieved by minimizing  $d(\hat y(x), \hat y(\tilde x))$ with respect to the trainable model  $\hat y$, where $d(, )$ measures a distance between the two outputs. In Equation \ref{eq:method}, these  methods can be viewed to control $V[f_{y}]$ by making the model $\hat y$ more invariant over the  space of $z$. Therefore, our theory formalizes the intuition of these regularization methods in terms of the generalization gap.  

In order to more effectively minimize the bound on the generalization gap in Equation \ref{eq:method}, we propose a new regularization method, called \textit{dual-cutout}. For each training input $x$, our dual-cutout method  minimizes the following   regularization loss $\ell_{\text{reg}}(x, \theta)$ with respect to $\theta$ (in addition to the original classification loss): 
$$
\ell_{\text{reg}}(x,\theta)= \int_{(x_{1},x_{2})} \|h(x_{1},\theta) - h(x_{2},\theta)\|_2 ^2 dP(x_{1}, x_{2}|x),
$$
 where $h(x',\theta)$ is the post-softmax output of the last layer of  a neural network with  parameters $\theta$ (given an input $x'$),  and $(x_{1}, x_{2}) \sim P(x_{1}, x_{2}|x)$ are two randomly sampled inputs of two random cutouts of a given natural input $x$. Here, we set $P(x_{1}, x_{2}|x)=P(x_{1}|x)P(x_{2}|x)$, and $P(x_{1}|x)=P(x_{2}|x)$ is the probability distribution over random cutout input $x_{1}$ given a original (non-cutout) input $x$; i.e., $P(x_{1}|x)=P(x_{2}|x)$ represents the same random cutout procedure as  single-cutout method  in the previous paper \citep{cutout}.  As this additional regularization loss gets smaller, the model becomes more insensitive over the  hidden space of $z$, implicitly minimizing $V[f_{y}]$ and the bound on the generalization gap in Equation \ref{eq:method}.

Table \ref{tbl:method} compares the test error of dual-cutout against single-cutout and the standard method for three benchmark datasets, namely CIFAR10, CIFAR100 and SVHN. Dual-cutout outperforms baseline methods by a significant margin.

\begin{table}[h!]
\centering
\caption{Test error (\%) with different regularization methods.   } \label{tbl:method}
\begin{tabular}{lcccc}
\toprule
Method & CIFAR-10 & CIFAR-100 & SVHN \\
\midrule
Standard        & 3.79 $\pm$ 0.07 & 19.85 $\pm$  0.14 & 2.47 $\pm$ 0.04 \\ 
\midrule
Single-cutout  & 3.19 $\pm$ 0.09 & 18.13 $\pm$ 0.28 &  2.23 $\pm$ 0.03  \\ 
\midrule
Dual-cutout    & 2.61 $\pm$ 0.04 & 17.54 $\pm$ 0.09 &  2.06 $\pm$ 0.06 \\ 
\bottomrule
\end{tabular} 
\end{table} 

We conducted all the experiments with the WideResNet$28\_10$ ~\citep{wrn} architecture and report the test errors at the end of   300 training epochs. We used SGD with the learning rate 0.1 and the momentum 0.9. At each step of SGD, to minimize the regularization loss $\ell_{\text{reg}}(x, \theta)$ of dual-cutout, we used the sampled gradient $\nabla_\theta \|h(x_{1}, \theta) - h(x_{2}, \theta)\|_2^2$ where $(x_{1}, x_{2})$ is sampled as $(x_{1}, x_{2}) \sim P(x_{1}, x_{2}|x)$. The learning rate was annealed at epochs 150 and 225 by a factor of 0.1. We used standard data-augmentation and preprocessing for all the datasets. For each dataset, we choose the  cutout size as reported in ~\citep{cutout}. We performed five trials of each experiment and report the standard deviation and mean of test error in Table ~\ref{tbl:method}.

\section{Discussion} \label{sec:beyond_classical_ML}

\begin{table*}[b!] \footnotesize
\centering
\caption{A simplified comparison, wherein GG denotes the generalization gap} \label{tbl:slt-vs-alt} 
\begin{tabular}{l | c | c}
\noalign{\hrule height 0.9pt}
& Statistical Learning Theory  & Analytical Learning Theory   \\  \noalign{\hrule height 0.4pt} 
GG is characterized by  & hypothesis spaces  $\mathcal H$ or algorithms $\mathcal A$  & a learned model $\hat y_{\mathcal A(S_{m})}$   \\ 
GG is decomposed via & statistics  & measure theory \\ 
Statistical assumption & is required  & can be additionally used \\ 
Main focus is when &  a (training) dataset $S_{m}$ remains random &  a (training) dataset $S_{m}$ is given
\\ 
Bounds on GG are  & not strongly instance-dependent & strongly instance-dependent\\
\noalign{\hrule height  0.9pt}
\end{tabular} 
\end{table*}

Table \ref{tbl:slt-vs-alt} summarizes the major simplified differences between statistical learning theory and  analytical learning theory. Because of the differences in the assumptions and the objectives, the proposed learning theory is not directly comparable in terms of sample complexity against previous learning theory. Instead of focusing on comparable sample-complexity, analytical learning theory focuses on complementing previous learning theory by providing additional practical insights. 
Indeed, the real-world phenomena that are analyzed are different in statistical learning theory and analytical learning theory.
Typically in statistical learning theory,  an upper bound holds  over  a fixed   $\mathcal H$ or  a fixed   $\mathcal A$  \textit{with high probability over different random datasets}. In contrast, in analytical learning theory, an upper bound holds \textit{individually for each problem instance}.

An another difference between  statistical learning theory and  analytical learning theory lies in the property of strong instance-dependence. Any generalization bound that depends on a non-singleton hypothesis space $\mathcal H \neq \{\allowbreak \hat y_{\mathcal A(S_{m})}  \}$, such as ones  with Rademacher complexity and VC dimension, is \textit{not} strongly instance-dependent because the non-singleton hypothesis space contains $\hat y \neq  \hat y_{\mathcal A(S_{m})}$, and the bound is not invariant  under an arbitrary change of $\mathcal H$. The definition of  stability  itself depends on  $\bar S_m$ that is not equal to $S_{m}$ and $ Z_{m'}$ \citep{bousquet2002stability}, making the corresponding bounds  be  \textit{not} strongly instance-dependent. Moreover, a generalization bound that depends on a concept of  random  datasets   $\bar S_m$ different from  $S_{m}$ and $ Z_{m'}$ (e.g.,  an additive term $O(\sqrt{1/m})$ that measures a deviation  from an expectation over  $\bar S_m\neq S_{m},Z_{m'}$) is  \textit{not} strongly instance-dependent, because the bound is not invariant under an arbitrary change of $\bar S_m$.

Data dependence  does not imply  strong instance-dependence.
For example, in the data-dependent bounds of the luckiness framework \citep{shawe1998structural,herbrich2002algorithmic},   the definition of $\omega$-smallness of the luckiness function  contains a non-singleton hypothesis space  $\mathcal H$, a  sequence of non-singleton hypothesis spaces  (ordered in a data-dependent way by a luckiness function), and a supremum over $\mathcal H$ with the probability over  datasets $\bar S_m \neq S_{m}$ (with $ Z_{m'}=S_{m})$ (e.g., see Definition 4 in \citealt{herbrich2002algorithmic} with contraposition). 
As exemplified in the luckiness framework,  one can usually turn   both data-dependent and data-independent bounds  into  more data-dependent ones by considering a sequence of hypothesis spaces or sets of learning algorithms. However, such data-dependent bounds still contain the complexity of a non-singleton hypothesis space (and  dependence on the definition of the sequence). The data-dependent bounds with empirical Rademacher complexity
\citep{koltchinskii2000rademacher,bartlett2002model} also depend on a non-singleton hypothesis space and its empirical Rademacher complexity. Moreover, the definition of robustness itself depends on   $\bar S_m$, which is not equal to $S_{m}$ or $ Z_{m'}$ 
\citep{xu2012robustness}. Therefore, all of these data-dependent bounds are not strongly instance-dependent.

The fact that Theorem \ref{thm:main}  is invariant to  the complexity of hypothesis space $\mathcal H$ and certain details of a learning algorithm $\mathcal A$ can be both advantageous and disadvantageous, depending on the objective of the analysis. As we move towards the goal of artificial intelligence, $\mathcal H$ and $\mathcal A$ would become extremely complex,  which can pose a challenge in theory. From this  viewpoint, analytical learning theory can also be  considered as a methodology to avoid such a challenge, producing  theoretical insights for intelligent systems with arbitrarily complex $\mathcal H$ and $\mathcal A$, so long as other conditions are imposed on the actual functions being computed by them.

\acks{We gratefully acknowledge support from NSF grants 1420316, 1523767 and 1723381,  from AFOSR FA9550-17-1-0165, from ONR grant N00014-14-1-0486, and from ARO grant  W911 NF1410433, as well as support from NSERC, CIFAR and Canada Research Chairs. Vikas Verma was supported by Academy of Finland project 13312683 / Raiko Tapani AT kulut. }

\bibliography{all}


\appendix

\begin{center}
\textbf{\Large
 Appendix
} \vspace{-0pt}
\end{center}

Appendix \ref{app:additional_explanation} contains additional discussions to facilitate understanding this paper. Appendix \ref{app:proofs} includes all the proofs of the theoretical results.   

\section{Additional discussions} \label{app:additional_explanation}

\subsection{An illustration of discrepancy} \label{app:illustration_discrepancy}
\begin{figure}[H] 
    \center
    \includegraphics[width=0.3\columnwidth]{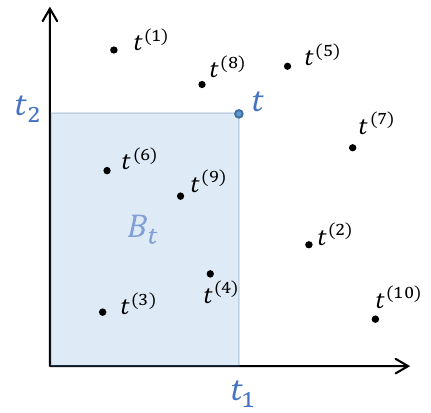}
    \caption{
       The local discrepancy $D[B_t;T_{m},\nu]$ evaluates the difference between the empirical measure of the box $B_t$ (the normalized number of data points in the box $B_t$, which is  $4/10$) and the measure $\nu$ of the box $B_t$ (the measure of the blue region)
    }  \label{fig:local_discrepancy}    
\end{figure}

Figure \ref{fig:local_discrepancy} shows an illustration of the  \textit{local discrepancy} $D[B_t;T_{m},\nu]$ and related notation in two dimensional space. 

\subsection{An illustration of a difference in the scopes of statistical and analytical learning theories} \label{app:illustration_difference_slt-alt}

\begin{figure}[b!]
\begin{subfigure}[b]{0.5\columnwidth}
  \includegraphics[width=\columnwidth]{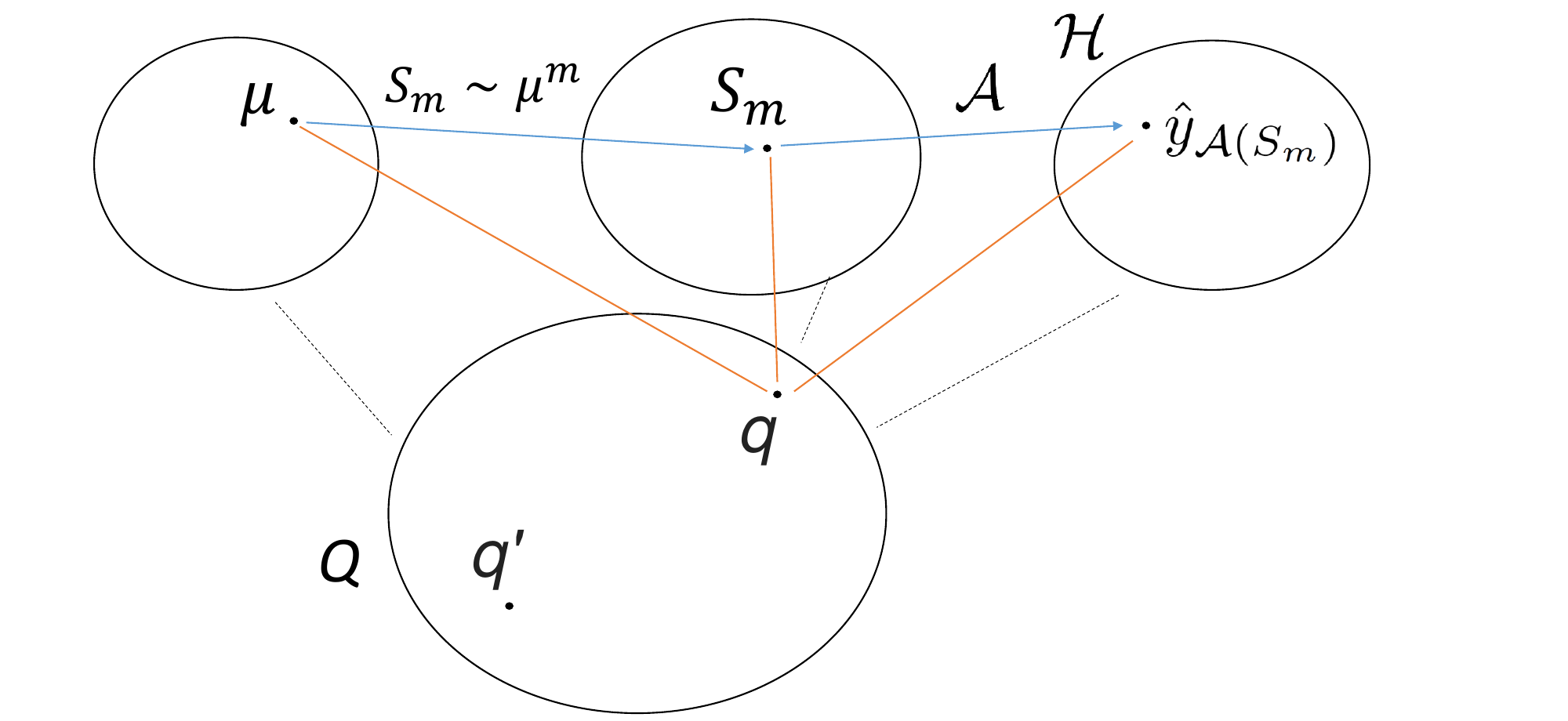}
  \caption{Statistical learning theory} 
\end{subfigure} 
\begin{subfigure}[b]{0.5\columnwidth}
  \includegraphics[width=\columnwidth]{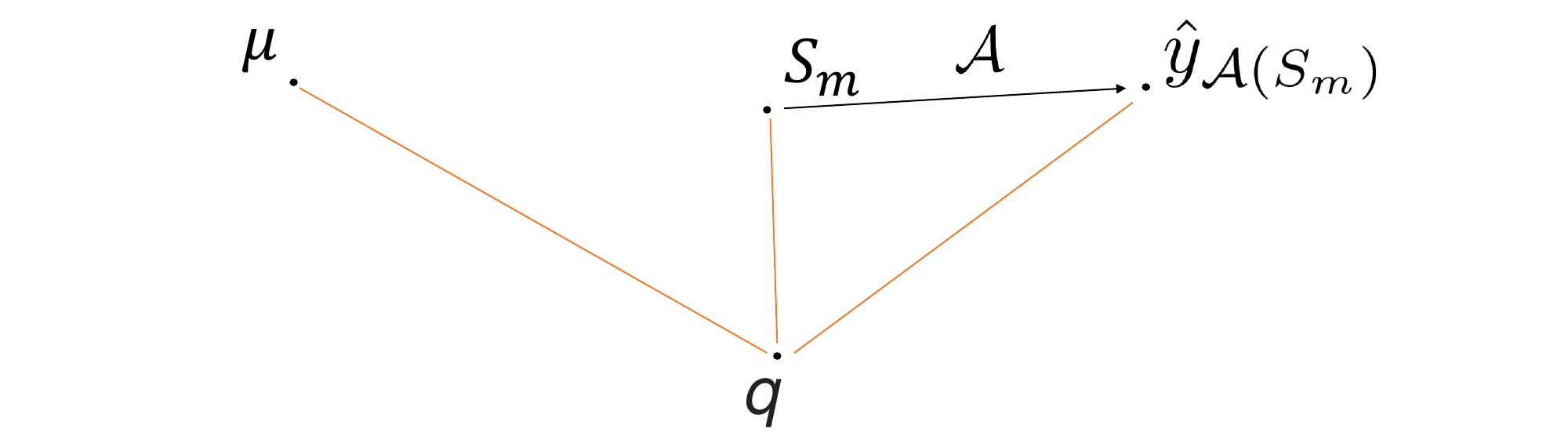}
  \caption{Analytical learning theory}
\end{subfigure}
\caption{An illustration of a difference in the scopes with $Z_{m'}=S_m$: $q$ represents a query about  the  generalization gap of a learned model $y_{\mathcal A(S_{m})}$, which is a deterministic quantity of the tuple $(\mu,S_{m},L\hat y_{\mathcal A(S_{m})})$. Intuitively, whereas analytical learning theory analyzes $q$ directly, statistical learning theory focuses more on analyzing the set $Q$ that contains $q$. The set $Q$ is defined by the sets of possible measures $\mu$ and randomly-drawn different datasets $S_{m}$ and the hypothesis space $\mathcal H$ or learning algorithm $\mathcal A$.}
\label{fig:scopes_slt-alt}
\end{figure}

Figure \ref{fig:scopes_slt-alt} shows a graphical illustration of a difference in the scopes of statistical learning theory and analytical learning theory. Here, $\mu^m$ is the product measure. 

In the setting of statistical learning theory (Figure \ref{fig:scopes_slt-alt} (a)), our typical goal is to analyze the random expected error $\EE_\mu[L  \hat y_{\mathcal A(S_{m})}]$ over the random datasets $S_{m}$ by fixing a hypothesis space and/or learning algorithm over random datasets. Due to the randomness over $S_{m}$, we do not know where $q$ exactly lands in $Q$. The lower bound and necessary condition in the setting of statistical learning theory is typically obtained via a worst-case instance $q'$ in $Q$.
For example, classical no free lunch theorems and lower bounds on the generalization gap via VC dimension (e.g., \citealt[Section 3.4]{mohri2012foundations}) have been derived with the worst-case distribution characterizing $q'$ in $Q$.  Such a necessary condition is only proven to be necessary for the worst-case $q' \in Q$, but is \textit{not} proven to be necessary for other ``good'' cases $q\neq q'$.  Intuitively, we are typically analyzing the quality of the set $Q$, instead of each  individual $q\in Q$. 

In this view, it becomes clear what is going on in some  empirical observations such as in \citep{zhang2016understanding}. Intuitively, whereas statistical learning theory focuses more on analyzing the set $Q$, each element such as $q$ (e.g., a ``good'' case or structured label case) and $q'$ (e.g., the worst-case or random label case) can significantly differ from each other.  Data-dependent analyses  in statistical learning theory can be viewed as the ways to decrease the size of $Q$ around each $q$. 

In contrast, analytical learning theory (Figure \ref{fig:scopes_slt-alt} (b)) focuses on each $q$ only, allowing tighter results for each ``good'' $q\in Q$ beyond the possibly ``bad'' quality of the set $Q$ overall.

It is important to note that analyzing the set $Q$ is  of  great interest on its own merits, and statistical learning theory has advantages over our proposed learning theory in this sense. Indeed, analyzing a set $Q$ is a natural task along the way of thinking in theoretical computer science (e.g., categorizing a set $Q$ of problem instances into polynomial solvable set or not). This situation  where theory focuses more on $Q$ and practical studies care  about each $q \in Q$ is prevalent in computer science even outside the learning theory. For example, the size of $Q$ analyzed in theory for optimal exploration in Markov decision processes (MDPs) has been  shown to be often too loose for each practical problem instance $q\in Q$, and a way to partially mitigate this issue was recently proposed \citep{kawaguchiAAAI2016}. Similarly, global optimization methods including Bayesian optimization approaches may suffer from  a large complex $Q$ for each practical problem instance $q\in Q$, which was partially mitigated in recent studies \citep{kawaguchiNIPS2015,kawaguchi2016global}. 

Furthermore, the issues of characterizing a set $Q$ only via a  worst-case instance $q'$ (i.e., worst-case analysis) are well-recognized in theoretical computer science, and so-called \textit{beyond worst-case analysis} (e.g., smoothed analysis) is an active research area to mitigate the issues.
Moreover,
a certain qualitative property of the set $Q$ might tightly capture that of each instance $q \in Q$. However, to prove such an assertion, proving that a upper bound on $\forall q \in Q$ matches a lower bound on $\exists q' \in Q$ is insufficient.

\subsection{On  usage of statistical assumption with $Z_{m'}=S_{m}$} \label{app:stat_assump}
Using a statistical assumption on a dataset $Z_{m'}$ with  $Z_{m'} \neq S_{m}$ is  consistent with a practical situation  where a  dataset $S_{m}$ is  given first instead of remaining random. For $Z_{m'} = S_{m}$, we can view this formulation as a mathematical modeling of the following situation. Consider  $S_{m}$ as a random variable  when collecting a dataset $S_{m}$, and then condition on the event of getting the collected dataset $S_{m}$ once $S_{m}$ is specified, focusing on minimization of  the (future) expected error $\EE_\mu[L  \hat y_{\mathcal A(S_{m})}]$ of the   model  $\hat y_{\mathcal A(S_{m})}$ \textit{learned with this particular specified dataset} $S_{m}$. 

In this  view, we can observe that if we draw an i.i.d.  dataset $S_{m}$, a   dataset $S_{m}$ is guaranteed to be statistically    ``good'' with high probability  in terms of  $D^*[\mathcal T_{*}\mu, \mathcal T(S_{m})]$ (e.g., $D^*[\mathcal T_{*}\mu, \mathcal T(S_{m})]\le c_{2} \sqrt{\frac{d}{m}}$    via Proposition \ref{prop:random_sample}). Thus, collecting a training dataset in a manner that satisfies the i.i.d. condition is  an effective method. However, once a  dataset $S_{m}$ is actually specified, there is no longer randomness over $S_{m}$, and the specified dataset $S_{m}$ is ``good'' (high probability event)  or ``bad'' (low probability event). We get   a  ``good'' dataset with high probability, and we obtain  probabilistic guarantees  such as Equation \eqref{eq:stat_assumption_0}. 

In many practical studies, a  dataset to learn a model is specified first as, for example, in studies with CIFAR-10, ImageNet, or  UCI datasets.
Thus, we might have a statistically  ``bad'' dataset  $S_{m}$ with no randomness over $S_{m}$  when these practical studies begin. Even then, we can  minimize the expected error in Theorem \ref{thm:main} by  minimizing $V[f]$ (and/or $D^*[\mathcal T_{*}\mu, \mathcal T(S_{m})]$ as deterministic quantity) such that  $V[f]D^*[\mathcal T_{*}\mu, \mathcal T(S_{m})]$ becomes marginal without the randomness over $S_{m}$.

\subsection{Supplementary explanation in Remark \ref{rem:fixed_feature_1}} \label{app:supp_rem_fixed_feature_1}
The bound is  always minimized if $\hat W=W^*$, but it is not a necessary condition in a pathological  case where the star-discrepancy $D^*$ is zero and $A_1$ can be zero with $\hat W\neq W^*$.

In Section \ref{sec:fixed_feature_true_model}, the optimal solution to minimize the expected error $\EE_{s} [\frac{1}{2}\|\hat W\phi(x)-y\|_2^2]$ is attained at  $\hat W=W^*$. To see this, we can expand the expected error as  
\begin{align*}
& \EE_{s}\left[\frac{1}{2} \|\hat W\phi (x)- y\|_2^2\right]  
\\ & =  \EE_x \left[ \frac{1}{2} \|\hat W\phi(x) -W^* \phi(x)\|_2^2 \right]  
+ \EE_{x,\xi} \left[\frac{1}{2}\|\xi\|_2^2 + \xi^\top \left(W^{*}\phi(x)-\hat W \phi(x) \right) \right]
\\ & =  \EE_x \left[ \frac{1}{2} \|\hat W\phi(x) -W^* \phi(x)\|_2^2 \right] + \EE_{\xi} \left[\frac{1}{2}\|\xi\|_2^2  \right], 
\end{align*} 
where the last line follows that $\xi$ is a zero-mean random variable independent of $x$. From the last line of the above equation, we can conclude the above statement about the minimizer.

\subsection{Flatness in  measures} \label{app:patho_non-flat}

It has been empirically observed that deep networks (particularly in the unsupervised setting) tend to transform the data distribution into  a flatter one closer to a uniform distribution in a space of a learned representation (e.g., see \citealt{bengio2013better}). If the  distribution $\mathcal T_*\mu$ with the learned representation $\mathcal T$ is uniform, then there exist better bounds on $ D^*[\mathcal T_*\mu,\mathcal T(Z_{m'})]$ such as $D^*[\mathcal T_*\mu,\mathcal T(Z_{m'})] \allowbreak \le10 \sqrt{d/m'}$ \citep{aistleitner2011covering}. Intuitively, if the measure $T_*\mu$  is non-flat and
concentrated near a highly curved manifold, then there are more opportunities for a greater mismatch between $\mathcal T_*\mu$ and $T(Z_{m'})$ to increase  $D^*[\mathcal T_*\mu,\mathcal T(Z_{m'})]$ (see below for pathological cases).  This  intuitively suggests the benefit of the flattening property that is sometimes observed with deep representation learning:  it is often illustrated with generative models or auto-encoders by showing how interpolating between the representations of two images (in representation space) corresponds (when projected in image space) to other images that are plausible (are on or near the manifold of natural images), rather than to the simple addition of two natural images \citep{bengio2009learning}.

If $\mathcal T_*\mu$ is concentrated in a single point, then $D^*[\mathcal T_*\mu,\mathcal T(Z_{m'})]=0$, but it implies that there is only a single  value of $L\hat y_{\mathcal A(S_{m})}(z)=f(\phi(x), v)$ because $(\phi(x), v)$ takes only one value. Hence, this is tight and consistent. On the other hand, to minimize the empirical error $\hat \EE_{Z_{m'}}[L\hat y_{\mathcal A(S_{m})}]$ with diverse label values,  $T_*\mu$ should not concentrate on the small number of finite points.

If  $ D^*[\mathcal T_*\mu,\mathcal T(Z_{m'})]$ is small, it means that the learned representation is effective at minimizing the generalization gap. This insight can be practically exploited by aiming to make $\mathcal T_*\mu$ flatter and spread out the data points $\mathcal T(Z_{m'})$ in a limited volume. It would also be beneficial to directly 
regularize an approximated $D^*[\mathcal T_*\mu,\mathcal T(Z_{m'})]$ with the unknown $\mu$ replaced by some known measures  (e.g., a finite-support measure corresponding to a validation dataset).

\subsection{Effect of higher-order derivatives} \label{app:high_deriv} 
 Example \ref{example:hidden}  suggests a method of regularization or model selection to control higher-order derivatives of a learned model w.r.t. a learned representation.  Let $f(t)= \ell(\hat Y (t), Y(t))$; here, $\hat Y$ and $Y$ represent the learned model $\hat y_{\mathcal A(S_{m})}$ and the target output $y$ as a function of $t=\mathcal T(z)$, respectively. Then, for example, if $\ell$ is the square loss, and if $\hat Y$ and $Y$ are smooth functions,  $V[f]$ goes to zero as  $\nabla^k \hat Y - \nabla^kY \rightarrow 0$ for $k=1,2,...,$ which can be upper bounded by $\|\nabla^k \hat Y\|+\|\nabla^kY\|$.

\section{Proofs} \label{app:proofs}
We use the following fact in our proof.  

\begin{lemma} \label{lemma:bounded_measurable} 
\emph{(theorem 3.1 in \citealt{aistleitner2017functions})}
Every real-valued function $f$ on $[0,1]^d$ such that $V[f]<\infty$ is Borel measurable.
\end{lemma}

\subsection{Proof of Proposition \ref{prop:variation_diff}}
\begin{proof}
By the definition, we have that 
\begin{align*}
\Delta^P_{j_{1},\dots,j_k} f_{j_{1}\dots j_k}(t_{j_1}^{(i_{1})}, \dots,t_{j_k}^{(i_{k})})
=\Delta^P_{j_{1},\dots,j_{k-1}} \left(\Delta^P_{j_{k}} f_{j_{1}\dots j_k}(t_{j_1}^{(i_{1})}, \dots,t_{j_k}^{(i_{k})}) \right)
\end{align*}
By the mean value theorem on the single  variable $t_{j_k}$, 
\begin{align*}
&\Delta^P_{j_{k}} f_{j_{1}\dots j_k}(t_{j_1}^{(i_{1})}, \dots,t_{j_k}^{(i_{k})})
= \left(\partial_{k}f_{j_{1}\dots j_k}(t_{j_1}^{(i_{1})}, \dots,c_{j_k}^{(i_{k})}) \right)(t_{j_k}^{(i_k+1)}-t_{j_k}^{(i_k)}),
\end{align*}
where $c_{j_k}^{(i_{k})} \in (t_{j_k}^{(i_{k}+1)},t_{j_k}^{(i_{k})})$. Thus, by repeatedly applying the mean value theorem, 
\begin{align*}
& \Delta^P_{j_{k}} f_{j_{1}\dots j_k}(t_{j      _1}^{(i_{1})}, \dots,t_{j_k}^{(i_{k})})
= \left(\partial_{1,\dots,k}^kf_{j_{1}\dots j_k}(c_{j_1}^{(i_{1})}, \dots,c_{j_k}^{(i_{k})}) \right) \prod_{l=1}^k (t_{j_k}^{(i_k+1)}-t_{j_k}^{(i_k)}),
\end{align*} 
where  $c_{j_l}^{(i_{l})} \in (t_{j_l}^{(i_{l}+1)},t_{j_l}^{(i_{l})})$ for all $l\in\{1,\dots,k\}$. Thus,
\begin{align*}
&V^{(k)}[f_{j_{1}\dots j_k}]
=\sup_{P \in \mathcal{P}_k} \sum_{i_1=1}^{m^P_1-1} \dots \sum_{i_{k}=1}^{ m^P_k-1} \left|\partial_{1,\dots,k}^kf_{j_{1}\dots j_k}(c_{j_1}^{(i_{1})}, \dots,c_{j_k}^{(i_{k})}) \right| 
 \prod_{l=1}^k (t_{j_k}^{(i_k+1)}-t_{j_k}^{(i_k)}).
\end{align*}
By taking supremum for $\left|\partial_{1,\dots,k}^kf_{j_{1}\dots j_k}(c_{j_1}^{(i_{1})}, \dots,c_{j_k}^{(i_{k})}) \right|$ and taking  it out from the sum, we obtain the first statement. The second statement follows the fact that if $\partial_{1,\dots,k}^kf_{j_{1}\dots j_k}(t_{j_1}^{(i_{1})}, \dots, \allowbreak t_{j_k}^{(i_{k})})$ is continuous, then $|\partial_{1,\dots,k}^kf_{j_{1}\dots j_k}(t_{j_1}^{(i_{1})}, \dots,t_{j_k}^{(i_{k})})|$ is continuous and Riemann  integrable. Thus, the right hand side on the above equation coincides with the definition of the Riemann integral of $|\partial_{1,\dots,k}^k f_{j_{1}\dots j_k}(t_{j_1}^{(i_{1})}, \dots,t_{j_k}^{(i_{k})})|$ over $[0,1]^k$. 
\end{proof}

\subsection{Proof of Theorem \ref{thm:main}}

The proof of Theorem \ref{thm:main} relies on several existing proofs from different fields. Accordingly, along the proof, we also track the extra dependencies and structures that appear only in machine learning, to confirm  the applicability of the previous proofs in the problem of machine learning.  Let $\one_{A}$ be an indicator function of a set $A$. Let $\Omega=[0,1]^d$. Let $\bold 1=(1,1,\dots,1)\in \Omega$ and $\bold 0=(0,0,\dots,0)\in \Omega$ as in a standard convention. The following lemma follows theorem 1.6.12 in \citep{ash2000probability}.

\begin{lemma} \label{lemma:right_level} 
For any $(\mathcal T,f) \in\mathcal{F}[L  \hat y_{\mathcal A(S_{m})}]$,
$$
\int_{\Zcal} f(\mathcal T(z)) d\mu(z) = \int_{\Omega} f(\omega) d(\mathcal T_{*}\mu)(\omega).
$$
\end{lemma}
\textit{Proof of Lemma \ref{lemma:right_level}.}
By Lemma \ref{lemma:bounded_measurable}, $f$ is a Borel measurable function. The rest of the proof of this lemma directly follows the proof of theorem 1.6.12 in \citep{ash2000probability}; we  proceed from simpler cases to more general cases as follows.
In the case of $f$ being an indicator function of some set $A$ as $f=\one_{A}$, we have that 
\begin{align*}
\int_{\Zcal} f(\mathcal T(z)) d\mu(z) &= \mu(\mathcal Z \cap \mathcal T^{-1} A) 
\\ &= (\mathcal T_{*}\mu)(\Omega\cap A)
\\ & = \int_{\Omega} f(\omega) d(\mathcal T_{*}\mu)(\omega). 
\end{align*}

In the case of  $f$ being a non-negative simple function as $f=\sum_{i=1}^n \alpha_i \one_{A_i}$,
\begin{align*}
\int_{\Zcal} f(\mathcal T(z)) d\mu(z) &= \sum_{i=1}^n \alpha_i \int_{\mathcal Z} \one_{A_i}(\mathcal T(z)) d\mu(z)
\\ & =\sum_{i=1}^n \alpha_i  \int_{\Omega} \one_{A_i}(\omega) d(\mathcal T_{*}\mu)(\omega)
\\ & = \int_{\Omega} f(\omega) d(\mathcal T_{*}\mu)(\omega),   
\end{align*}
where the second line follows what we have proved for the case of $f$ being an indicator function. 

In the case of $f$ being a non-negative
Borel measurable function, let $(f_k)_{k\in \NN}$ be an increasing sequence of simple functions such that $f(\omega)=\lim_{k \rightarrow \infty} f_k(\omega)$, $\omega \in \Omega$. Then, by what we have proved for simple functions, we have $\int_{\Zcal} f_{k}(\mathcal T(z)) d\mu(z) = \int_{\Omega} f_{k}(\omega) d(\mathcal T_{*}\mu)(\omega)$. Then, by the monotone convergence theorem,
we have $\int_{\Zcal} f(\mathcal T(z)) d\mu(z) = \int_{\Omega} f(\omega) d(\mathcal T_{*}\mu)(\omega)$.

In the case of $f=f^{+}-f^-$ being an arbitrary Borel measurable function,
we have already proved  the desired statement for each $f^{+}$ and $f^{-}$, and by the definition of Lebesgue integration, the statement for $f$ holds.

$\hfill \square$

\textit{Proof of Theorem \ref{thm:main}.} With Lemmas \ref{lemma:bounded_measurable} and \ref{lemma:right_level},  the proof follows that of theorem 1 in \citep{aistleitner2015functions}. For any $(\mathcal T,f) \in \mathcal F[L\hat y_{\mathcal A(S_{m})}]$,
 \begin{align*}
 \int_{\Zcal} L\hat y_{\mathcal A(S_{m})}(z) d\mu(z)- \frac{1}{{m'}}\sum_{i=1}^{m'} L\hat y_{\mathcal A(S_{m})}(z_i) 
 & = \int_{\Zcal} f(\mathcal T(z)) d\mu(z)- \frac{1}{{m'}}\sum_{i=1}^{m'} f(\mathcal T(z_{i})) 
\\ & = \int_{\Omega} f(\omega) d(\mathcal T_{*}\mu)(\omega)- \frac{1}{{m'}}\sum_{i=1}^{m'} f(\mathcal T(z_{i})) 
\end{align*}
where the second line follows the condition of $\mathcal T$ and $f$ and the third line follows Lemma \ref{lemma:right_level}. In the following, we first
consider the case where $f$ is left-continuous, and then
discard the left-continuity condition later.

\uline{Consider the case where $f$ is left-continuous (for the second statement):} Suppose that $f$ is left-continuous coordinate-wise at every point in the domain. Given a pair of vectors  $(a,b)$, we write $a\le b$ if  the relation holds for every coordinate.  Let $\tilde f(\omega)=f(\bold 1-\omega)-f(\bold 1)$ for all $\omega \in \Omega$. Then, by theorem 3 and equation (20) in \citep{aistleitner2015functions}, there exists signed Borel measure $\mu_{\tilde f}$ on $\Omega$ such that ${\tilde f}(\omega)=\mu_{\tilde f}([\bold 0,\omega])$ for all $\omega \in \Omega$ and $|\mu_{\tilde f}|(\Omega)=V[f]+|\tilde f(0)|=V[f]$. Let $\nu_{f}$ be the reflected measure of $\mu_{\tilde f}$ as $\nu_{ f}(A)=\mu_{\tilde f}(\bold 1 - A)$ for any Borel set $A \subset \Omega$ where $\bold 1 - A=\{\bold 1 - t: t \in A\}$. It follows that $\nu_{f}$ is a signed Borel measure and 
\begin{displaymath}
|\nu_{f}|(\Omega)=|\mu_{\tilde f}|(\Omega)=V[f].
\end{displaymath}    
By using these, we can rewrite $f$ as 
\begin{align*}
f(\omega) &= f(\bold 1)+{\tilde f}(\bold 1-\omega) 
\\ &=f(\bold 1)+ \int_\Omega \one_{[\bold 0,\bold 1-\omega]}(t) d\mu_{\tilde f}(t)
\\ &=f(\bold 1)+ \int_\Omega \one_{[\omega,\bold 1]}(t) d \nu_{f}(t)
\\ & =f(\bold 1)+ \int_\Omega \one_{[\bold 0,t]}(\omega) d \nu_{f}(t), 
\end{align*}
where the second line follows from $\{\bold 1 - t: t \in [\omega,\bold 1]\}=[\bold 0,\bold 1-\omega]$. Then, by linearity, 
$$
\frac{1}{{m'}}\sum_{i=1}^{m'} f(\mathcal T(z_{i})) - f(\bold 1)= \int_\Omega \frac{1}{{m'}}\sum_{i=1}^{m'}\one_{[\bold 0,t]}(\mathcal T(z_{i})) d \nu_{f}(t),
$$
and by 
the Fubini--Tonelli theorem and linearity, 
\begin{align*}
\int_{\Omega} f(\omega) d(\mathcal T_{*}\mu)(\omega)- f(\bold 1)
 &= \int_\Omega  \int_\Omega \one_{[\bold 0,t]}(\omega)  d(\mathcal T_{*}\mu)(\omega)d \nu_{f}(t)
\\ &= \int_\Omega  (\mathcal T_{*}\mu)([\bold 0,t]) d \nu_{f}(t).
\end{align*}
Therefore,
\begin{align*}
&\int_{\Omega} f(\omega) d(\mathcal T_{*}\mu)(\omega)- \frac{1}{{m'}}\sum_{i=1}^{m'} f(\mathcal T(z_{i})) 
= \int_\Omega \left( (\mathcal T_{*}\mu)([\bold 0,t])-\frac{1}{{m'}}\sum_{i=1}^{m'}\one_{[\bold 0,t]}(\mathcal T(z_{i})) \right) d \nu_{f}(t),
\end{align*}
which  proves the second statement of this theorem by noticing that $f(t)= \nu_{f}([t,\bold 1])+f(\bold 1)$. Moreover,
this implies that
\begin{align*}
\left| \int_{\Omega} f(\omega) d(\mathcal T_{*}\mu)(\omega)- \frac{1}{{m'}}\sum_{i=1}^{m'} f(\mathcal T(z_{i})) \right|
 & \le  |d \nu_{f}(t) |(\Omega)D^*[\mathcal T_{*}\mu,\mathcal T(Z_{m'})]
\\ & =V[f] D^*[\mathcal T_{*}\mu,\mathcal T(Z_{m'})].
\end{align*}

\uline{Discard the left-continuity condition of $f$ (for the first statement):} Let $f$ be given and fixed without left-continuity condition. For each fixed $f$, by
the law of large numbers (strong law of large numbers
and the multidimensional Glivenko--Cantelli theorem),
for any $\epsilon>0$, there exists a number $n$ and a set $\bar A_{n}=\{\bar \omega_i\}_{i=1}^n$ such that  both of the following two inequalities hold:
$$
\left| \int_{\Omega} f(\omega) d(\mathcal T_{*}\mu)(\omega) - \frac{1}{n}\sum_{i=1}^n f(\bar \omega_i) \right| \le \epsilon,
$$
and
$$
D^*[\mathcal T_{*}\mu,\bar A_n] \le \epsilon.
$$
Let $\bar A_{n}=\{\bar \omega_i\}_{i=1}$ be such a set. For each fixed $f $,  let  $f_n$ be a left-continuous function such that  $f_{n} (\omega)=f(\omega)$ for all $\omega \in \bar A_{n} \cup \mathcal T(Z_{m'})$ and $V[f_{n}]\le V[f]$. This definition of $f_n$ is non-vacuous and we can  construct such a $f_n$ as follows. Let $\mathcal G$ be the $d$-dimensional grid generated by the set $\{\bold 0\} \cup \{\bold 1\} \cup \bar A_{n} \cup \mathcal T(Z_{m'})$; $\mathcal G$ is the set of all points $\omega \in \Omega$ such that for $k\in \{1,\dots,d\}$, the $k$-th coordinate value of $\omega$ is the $k$-th coordinate value of some element in the set $\{\bold 0\} \cup \{\bold 1\} \cup \bar A_{n} \cup \mathcal T(Z_{m'})$. We  can construct a desired $f_n$ by setting
$
f_{n}(\omega)=f(\text{succ}_{n}(\omega)),
$ 
where $\text{succ}_{n}(\omega)$ outputs an unique element $t \in \mathcal G$ satisfying the condition that $t \ge \omega$ and $t \le t'$ for all $t' \in \{t'\in \mathcal G:t' \ge \omega \}$.

Then, by triangle inequality, we write  
\begin{align*}
\left|\int_{\Omega} f(\omega) d(\mathcal T_{*}\mu)(\omega)- \frac{1}{{m'}}\sum_{i=1}^{m'} \underbrace{f(\mathcal T(z_{i}))}_{=f_{n} (\mathcal T(z_{i}))}\right|
 & \le \left|\int_{\Omega} f_n(\omega) d(\mathcal T_{*}\mu)(\omega)- \frac{1}{{m'}}\sum_{i=1}^{m'} f_n(\mathcal T(z_{i}))\right|
\\ & \hspace{10pt} +  \left|\frac{1}{n}\sum_{i=1}^n \underbrace{f_n (\bar \omega_i)}_{=f(\bar \omega_i)} - \int_{\Omega} f_n(\omega) d(\mathcal T_{*}\mu)(\omega)\right| \\ & \hspace{10pt} + \left| \int_{\Omega} f(\omega) d(\mathcal T_{*}\mu)(\omega) - \frac{1}{n}\sum_{i=1}^n f (\bar \omega_{i}) \right|.
\end{align*} 
Because $f_n$ is left-continuous, we can apply our previous result to the first and the second terms;   the first term is at most $V[f_n]D^*[\mathcal T_{*}\mu,T(Z_{m'})]\le  V[f]D^*[\mathcal T_{*}\mu,T(Z_{m'})]$, and the second term is at most $V_{}[f_n]D^*[\mathcal T_{*}\mu,\bar A_{n}]\le \epsilon V_{}[f]$. The third term is at most $\epsilon$ by the definition of $\bar A_n$. Since $\epsilon>0$ can be arbitrarily small, we have that for each  $(f,\mathcal T) \in \mathcal F[L\hat y_{\mathcal A(S_{m})}]$, (deterministically,)
\begin{align*}
&\left| \int_{\Omega} f(\omega) d(\mathcal T_{*}\mu)(\omega)- \frac{1}{{m'}}\sum_{i=1}^{m'} f(\mathcal T(z_{i})) \right|
\le V[f]D^*[\mathcal T_{*}\mu,T(Z_{m'})].
\end{align*}

\uline{Putting together:} for any $(\mathcal T,f) \in \mathcal F[L\hat y_{\mathcal A(S_{m})}]$, 
\begin{align*}
& \left|\int_{\Zcal} L\hat y_{\mathcal A(S_{m})}(z) d\mu(z)- \frac{1}{{m'}}\sum_{i=1}^{m'} L\hat y_{\mathcal A(S_{m})}(z_i)  \right| 
 \le V[f] D^*[\mathcal T_{*}\mu,\mathcal T(Z_{m'})] 
\end{align*}
Thus, $\left|\int_{\Zcal} L\hat y_{\mathcal A(S_{m})}(z) d\mu(z)- \frac{1}{{m'}}\sum_{i=1}^{m'} L\hat y_{\mathcal A(S_{m})}(z_i)  \right|$ is a lower bound of a set $Q =\{V[f]  \allowbreak D^*[\mathcal T_{*}\mu, \allowbreak\mathcal T(Z_{m'})] :(\mathcal T,f) \in \mathcal{F}[L  \hat y_{\mathcal A(S_{m})}]\}$. By the definition of infimum, $|\int_{\Zcal} L\hat y_{\mathcal A(S_{m})}(z)  d\mu(z) \allowbreak  - \frac{1}{m'}\sum_{i=1}^{m'} \allowbreak L\hat y_{\mathcal A(S_{m})}(z_i)  |  \le  \inf Q$, if $\inf Q$ exists. Because $Q$ is a nonempty subset of real and lower bounded by $0$, $\inf Q$ exists. Therefore,
\begin{align*}
&\left| \int_{\Zcal} L\hat y_{\mathcal A(S_{m})}(z) d\mu(z)- \frac{1}{{m'}}\sum_{i=1}^{m'} L\hat y_{\mathcal A(S_{m})}(z_i) \right| 
 \le \inf_{(\mathcal T,f) \in \mathcal{F}[L  \hat y_{\mathcal A(S_{m})}]} V[f] D^*[\mathcal T_{*}\mu, \mathcal T(Z_{m'})],
\end{align*}
which implies the  first statement of this theorem.

$\hfill \square$

\subsection{Proof of Proposition \ref{prop:random_sample}}
\begin{proof}
From theorem 2 in \citep{heinrich2001inverse}, there exists a positive constant $c_{1}$ such that for all $s \ge c_{1} \sqrt d$ and for all ${m'} \in \NN^+$,
$$
\PP \left\{D^*[\mathcal T_{*}\mu, \mathcal T(Z_{m'})] \ge s {m'}^{-1/2} \right\}  \le \frac{1}{s} \left(\frac{c_{1} s^2}{d} \right)^{d} e^{-2s^{2}},
$$
where we used the fact that the VC dimension of the set  of the axis-parallel boxes contained in $[0,1]^d$ with one vertex at the origin is $d$ (e.g., see \citealt{dudley1984course}). By setting $s=c_2\sqrt d$ for any $c_2 \ge c_1$, we obtain the desired result. 
\end{proof}

\subsection{Proof of Proposition \ref{prop:determ_sample}}
\begin{proof}
From theorem 1 in \citep{aistleitner2014low}, for any ${m'} \in \NN^+$, there exists a set $T_{m'}$ of points $t_1,\dots,t_{m'} \in [0,1]^d$ such that 
$$
D^*[\mathcal T_{*}\mu, T_{m'}] \le 63 \sqrt d \frac{(2+\log_2 {m'})^{(3d+1)/2}}{{m'}}.
$$
Because $\mathcal T$ is a surjection, for such a $T_{m'}$, there exists $Z_{m'}$ such that $\mathcal T(Z_{m'})= T_{m'}$. 
\end{proof}

\subsection{Proof of the inequality in Example \ref{example:loss}} \label{app:proof_gexample-loss}
Let $\mu_{\mathcal T(Z_{m'})}$ be
a (empirical) normalized measure with the finite support on $\mathcal T(Z_{m'})$. Then,  
\begin{align*}
\EE_{\mu}[L  \hat y_{\mathcal A(S_{m})}] -  \hat \EE_{Z_{m'}}[L  \hat y_{\mathcal A(S_{m})}] 
 &\le V[f] D^*[\mathcal T_{*}\mu, \mathcal T(Z_{m'})]
\\ &=\max\{|(\mathcal T_{*}\mu)(\{0\})- \mu_{\mathcal T(Z_{m'})}(\{0\})|,
\\  & \hspace{37pt}     |(\mathcal T_{*}\mu(\{0,1\})) - \mu_{\mathcal T(Z_{m'})}(\{0,1\})|\} \\ & =|\mathcal T_{*}\mu(\{0\})-\mathcal  \mu_{\mathcal T(Z_{m'})}(\{0\})|
\\ & =|1-\mathcal T_{*}\mu(\{1\})- 1 +\mathcal  \mu_{\mathcal T(Z_{m'})}(\{1\})| 
\\ & =|\mathcal T_{*}\mu(\{1\})-\mathcal \mu_{\mathcal T(Z_{m'})}(\{1\})|.
\end{align*} 
Rewriting $\mu_{\mathcal T(Z_{m'})}(\{1\})=\EE_{Z_{m'}}[L  \hat y_{\mathcal A(S_{m})}]$ yields the desired inequality in Example \ref{example:loss}. 

\subsection{Proof of Theorem \ref{thm:fixed_feature_1}}
\begin{proof}  Let $L\hat y_{\mathcal A(S_{m})}(x)=\frac{1}{2} \|\hat W\phi(x) -W^* \phi(x)\|_2^{2}$ ($\Zcal = \Xcal$). Since 
\begin{align*}
\frac{1}{2} \|W\phi (x)- y\|_2^2  = &\frac{1}{2} \|W\phi(x) -W^* \phi(x)\|_2^2 
 + \frac{1}{2}\|\xi\|_2^2 - \xi^\top \left(W\phi(x)-W ^{*}\phi(x) \right) ,
\end{align*}
we have 
\begin{align*}
&\EE_s\left[\frac{1}{2}\|\hat W\phi (x)- y\|_2^2\right] - \hat \EE_{S_{m}} \left[\frac{1}{2}\|\hat W\phi (x)- y\|_2^2\right]
\\ & =\EE_{\mu_x} [L\hat y_{\mathcal A(S_{m})}]- \hat \EE_{X_m} [L\hat y_{\mathcal A(S_{m})}] + A_1 + A_2
\\ & \le V[f]D^{*}[\phi_*\mu_x, \phi(X_{m})] + A_1 + A_2 , 
\end{align*}

where 
the last line is obtained by applying Theorem \ref{thm:main} to $\EE_{\mu_x} [L\hat y_{\mathcal A(S_{m})}]- \hat \EE_{X_m} [L\hat y_{\mathcal A(S_{m})}]$ as follows. Let  $\mathcal T(x) = \phi(x)$ and  $f(t)=\frac{1}{2}\|\hat Wt-W^* t\|_2^{2}$, where $t\in \RR^{d_\phi}$. Then,  $L\hat y_{A(S_{m})}(x)=(f \circ \mathcal T)(x)$, and $(\mathcal T,f) \in \mathcal{F}[L  \hat y_{\mathcal A(S_{m})}]$ in Theorem \ref{thm:main} if $V[f]<\infty$. Therefore, by Theorem \ref{thm:main}, if $V[f]<\infty$,
\begin{displaymath}
\EE_{\mu_x} [L\hat y_{\mathcal A(S_{m})}]-  \hat \EE_{X_m} [L\hat y_{\mathcal A(S_{m})}]  \le V[f]D^{*}[\mathcal \phi_*\mu_x,\phi(X_{m})]. 
\end{displaymath}
To upper bound $V[f]$ and to show $V[f]<\infty$, we invoke Proposition \ref{prop:variation_diff} as follows. We have that $\frac{\partial f}{\partial t_l}= (\hat W_{ l} - W_{ l}^*)^\top (\hat W - W^*)t$, and  $\frac{\partial f}{\partial t_l \partial t_{l'}}= (\hat W_{ l} - W_{ l}^*)^\top (\hat W_{ l'} - W_{ l'}^*)$.
Because the second derivatives are constant over  $t$, the third and higher derivatives are zeros. Let $\tilde t_l=(t_{1},\dots,t_{d_\phi})^\top$ with $t_j\equiv1$ for all $j \neq l$. 
Then, we have that 
\begin{align*}
\sum_{l=1}^{d}V^{(1)}[f_{l}]
&=\sum_{l=1}^d \int_{[0,1]} |(\hat W_{ l} - W_{ l}^*)^\top (\hat W - W^*)\tilde t_l |dt_l
\\ & \le \sum_{l=1}^d \|(\hat W_{ l} - W_{ l}^*)^\top (\hat W - W^*)\|_{1} \int_{[0,1]}  \|\tilde t_l\|_\infty dt_l.
\\ & = \sum_{l=1}^d \|(\hat W_{ l} - W_{ l}^*)^\top (\hat W - W^*)\|_1,  
\end{align*}
and
\begin{align*}
\sum_{1\le l < l' \le d} V^{(2)}[f_{ll'}] 
& \le \sum_{1\le l < l' \le d} |(\hat W_{ l} - W_{ l}^*)^\top (\hat W_{ l'} - W_{ l'}^*)|.
\end{align*} 
Since higher derivatives exist and are zeros, from Proposition \ref{prop:variation_diff}, $V^{(k)}[f_{j_{1}\dots j_k}]=0$ for $k=3,\dots,d$. By the definition of $V[f]$, we obtain the desired bound for $V[f]$, and we have $V[f]<\infty$ if $\|\hat W - W^*\|<\infty$ (where there is no need to specify the particular matrix norm because of the equivalence of the norm). 
\end{proof}

\subsection{Proof of Theorem \ref{thm:fixed_feature_2}}

\begin{proof}
Let $W_{l' l}$ be the $(l',l)$-th entry of the matrix   $W$. Let $L\hat y_{\mathcal A(S_{m})}(s) = \frac{1}{2}\|\hat W\phi(x)-y\|_2^2$ ($\Zcal = \Xcal \times \Ycal$). Let    
$\mathcal T(s) = (\phi(x),y)$ and  $f(t,y)=\frac{1}{2}\|\hat Wt-y\|_2^{2}$ . Then,  $\ell(s)=(f \circ \mathcal T)(s)$, and $(\mathcal T,f) \in \mathcal{F}[L  \hat y_{\mathcal A(S_{m})}]$ in Theorem \ref{thm:main} if $V[f]<\infty$. Therefore, by Theorem \ref{thm:main}, if $V[f]<\infty$,
\begin{align*}
&\EE_s\left[\frac{1}{2}\|\hat W\phi (x)- y\|_2^2\right] - \hat \EE_{S_{m}} \left[\frac{1}{2}\|\hat W\phi (x)- y\|_2^2\right]
 \le V[f]D^{*}[\mathcal T_*\mu_{s}, \mathcal T(S_{m})]. 
\end{align*}

To upper bound $V[f]$ and to show $V[f]<\infty$, we invoke Proposition \ref{prop:variation_diff} as follows. For the first derivatives, we have
that $\frac{\partial f}{\partial t_l}= \hat W^\top_{ l} (\hat Wt - y)$ and $\frac{\partial f}{\partial y_l}= - (\hat Wt - y)_{l}$.
For the second derivatives, we have that  
$\frac{\partial^2 f}{\partial t_l \partial t_{l'}} = \hat W^\top_{ l} \hat W_{ l'}$, 
\begin{displaymath}
\frac{\partial ^{2}f}{\partial y_l \partial y_{l'}} =
\begin{cases}1 & \text{if } l=l' \\
0 & \text{if } l \neq l,
\end{cases}
\end{displaymath}
and
 $\frac{\partial^2 f}{\partial t_l \partial y_{l'}} = -\hat W_{l' l}$.  Because the second derivatives are constant in  $t$ and $y$, the third and higher derivatives are zeros. Then, because $|\frac{\partial f}{\partial t_l}| \le M \|\hat W_{ l}\|_1$ and $|\frac{\partial f}{\partial y_l}|\le M$, with $l=j_1$,
\begin{displaymath}
\sum_{j_{1}=1}^{d_\phi}V^{(1)}[f_{j_{1}}] \le M \sum_{l=1}^{d_\phi}\|\hat W_{l}\|_1, 
\end{displaymath}
and
\begin{displaymath}
\sum_{j_{1}=d_{\phi}+1}^{d_\phi+d_y}V^{(1)}[f_{j_{1}}] \le d_{y}M.  
\end{displaymath}
Furthermore, 
for $j_{1},j_{2} \in \{1,\dots,d_\phi\}$,
with $l=j_1$ and $l'=j_2$,
\begin{displaymath}
V^{(2)}[f_{j_{1}j_{2}}] \le |\hat W^\top_{l} \hat W_{l'}|. 
\end{displaymath}
For $j_1 \in \{1,\dots,d_\phi\}$ and $j_2\in \{d_\phi+1,\dots,d_\phi+d_y\}$, with $l=j_1$ and $l'=j_2-d_\phi$,
\begin{displaymath}
V^{(2)}[f_{j_1j_2}] \le |\hat W_{l'l}|,
\end{displaymath}
and for $j_{1},j_{2}\in \{d_\phi+1,\dots,d_\phi+d_y\}$,  
\begin{displaymath}
V^{(2)}[f_{j_{1}j_{2}}] \le 
\begin{cases}
1 & \text{ if } j_{1}=j_{2} \\
0 & \text{ otherwise}.
\end{cases}
\end{displaymath}

Thus,
\begin{align*}
\sum_{1\le j_1 < j_2 \le d_\phi+d_y} V^{(2)}[f_{j_{1}j_2}]
& = \sum_{1\le l<l' \le d_\phi} |\hat W^\top_{ l} \hat W_{ l'}| +\sum_{l=1} ^{d_\phi} \sum_{l'=1}^{d_{y}}|\hat W_{l' l}| 
\\ & = \sum_{1\le l<l' \le d_\phi} |\hat W^\top_{ l} \hat W_{ l'}| + \sum_{l=1}^{d_\phi}\|\hat W_{ l}\|_1.
\end{align*}

Therefore,
\begin{align*}
V[f] 
& = \sum_{k=1}^{d_\phi+d_y}  \ \ \sum_{1\le j_1 <  \dots < j_k \le d_\phi+d_y} V^{(k)}[f_{j_{1}\dots j_k}] 
\\ & = \sum_{k=1}^2  \ \ \sum_{1\le j_1 <  \dots < j_k \le d_\phi+d_y} V^{(k)}[f_{j_{1}\dots j_k}]
\\ & \le (M+1) \sum_{l=1}^{d_\phi}\|\hat W_{ l}\|_1 +\sum_{1\le l<l' \le d_\phi} |\hat W^\top_{ l} \hat W_{ l'}|+ d_{y} M.
\end{align*}
Here, we have $V[f]<\infty$ because $\|\hat W\|<\infty$ and $M<\infty$ (and the equivalence of the norm).
\end{proof}

\end{document}